\documentclass{article}

\usepackage[preprint]{neurips_2024}

\usepackage[utf8]{inputenc} 
\usepackage[T1]{fontenc}    
\usepackage{hyperref}       
\usepackage{url}            
\usepackage{booktabs}       
\usepackage{amsfonts}       
\usepackage{nicefrac}       
\usepackage{microtype}      
\usepackage{xcolor}         
\usepackage{graphicx}
\usepackage{subfigure}
\usepackage{wrapfig}
\usepackage{amsmath}
\usepackage{multirow}
\usepackage{arydshln}
\usepackage{algorithm}
\usepackage{algpseudocode}

\usepackage{amssymb}
\newtheorem{theorem}{Theorem}[section]
\newtheorem{proposition}[theorem]{Proposition}
\newtheorem{lemma}[theorem]{Lemma}

\newtheorem{assumption}[theorem]{Assumption}

\usepackage{hyperref}
\title{
1-Bit FQT: Pushing the Limit of Fully Quantized Training to 1-bit
}

\author{%
  Chang Gao \\
  Beijing Jiaotong University\\
  \texttt{22110098@bjtu.edu.cn} \\
   \And
   Jianfei Chen \\
   Tsinghua University \\
   \texttt{jianfeic@tsinghua.edu.cn} \\
   \AND
   Kang Zhao \\
   Tsinghua University \\
   \texttt{zhaokang29@huawei.com} \\
   \And
   Jiaqi Wang \\
   Beijing Jiaotong University \\
   \texttt{jiaqi.wang@bjtu.edu.cn} \\
   \And
   Liping Jing \\
   Beijing Jiaotong University \\
   \texttt{lpjing@bjtu.edu.cn} \\
}

\begin{document}

\maketitle

\begin{abstract}
  Fully quantized training (FQT) accelerates the training of deep neural networks by quantizing the activations, weights, and gradients into lower precision. To explore the ultimate limit of FQT (the lowest achievable precision), we make a first attempt to 1-bit FQT. We provide a theoretical analysis of FQT based on Adam and SGD, revealing that the gradient variance influences the convergence of FQT. Building on these theoretical results, we introduce an Activation Gradient Pruning (AGP) strategy. The strategy leverages the heterogeneity of gradients by pruning less informative gradients  and enhancing the numerical precision of remaining gradients to mitigate gradient variance. Additionally, we propose Sample Channel joint Quantization (SCQ), which utilizes different quantization strategies in the computation of weight gradients and activation gradients to ensure that the method is friendly to low-bitwidth hardware. Finally, we present a framework to deploy our algorithm. For fine-tuning VGGNet-16 and ResNet-18 on multiple datasets, our algorithm achieves an average accuracy improvement of approximately 6\%, compared to per-sample quantization. Moreover, our training speedup can reach a maximum of 5.13× compared to full precision training. Ours code is available at \href{https://github.com/Gaochang-bjtu/1-bit-FQT.git}{https://github.com/Gaochang-bjtu/1-bit-FQT}.
\end{abstract}

\section{Introduction} \label{introduction}

Training neural networks has a high computational cost and memory footprint. Training with low-precision arithmetic (a.k.a., fully quantized training or FQT) can enhance computational and memory efficiency. 
FQT quantizes weights, activations, and gradients into low-bitwidth numerical formats, enabling a fast implementation of both forward and backward propagation on low-precision hardware. 

The speedup potential of FQT depends on the numerical precision. Research aims to reduce the training numerical precision, without compromising convergence speed or accuracy. The required precision has been reduced from FP/INT16 \cite{micikevicius2017mixed,das2018mixed} to FP/INT8 \cite{wang2018training,banner2018scalable,zhu2020towards,yang2020training}. As of now, some work  \cite{sun2020ultra,chmiel2021logarithmic,xi2023training} have successfully pushed precision down to 4 bits.\par

As the training numerical precision continues to decrease, a natural question arises: 
\begin{center}
\emph{What is the ultimate limit of FQT (i.e., the minimum achievable bitwidth)? }
\end{center}
Answering this question not only advances our understanding of FQT but also provides a crucial direction for future hardware design strategies. 
Ideally, if we can push the bitwidth down to 1-bit, the training can be implemented with binary operations, such as XNOR and bitcounting operations \cite{courbariaux2016binarized}, and hardware design might be greatly simplified. Binary computation is already shown possible for \emph{inference} acceleration, such as XNOR-Net \cite{rastegari2016xnor}, but 1-bit \emph{training}  remains unexplored.

\begin{wrapfigure}[15]{r}{0.43\textwidth}
  \begin{center}
    \includegraphics[width=0.42\textwidth]{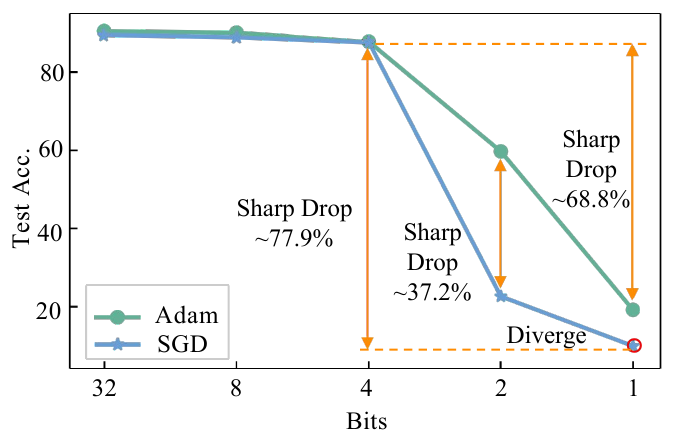}
  \end{center}
  \vspace{-0.2in}
  \caption{Gradient numerical precision (``bits'') vs. test accuracy of VGGNet16 on CIFAR-10, trained with Adam and SGD.}
  \vspace{-0.1in}
  \label{Adam-SGD}
\end{wrapfigure}

Reducing the bitwidth for FQT is challenging because of (1) the lack of theoretical understanding, especially how gradient quantization affects the convergence; (2) the large quantization error of gradients, which causes a sharp performance drop or even divergence when reducing gradient bitwidth lower than 4-bit (Fig.~\ref{Adam-SGD}). Due to these challenges,  current research frontier is still 4-bit FQT.

In this work, we make a first attempt towards achieving 1-bit FQT. Firstly, we provide a theoretical analysis for FQT based on both Adam \cite{kingma2014adam} and SGD. Our analysis links the convergence with gradient variance. Specifically, our analysis reveals that Adam is more suitable for FQT than SGD in the low-bitwidth regime, due to their different sensitivity to gradient variance. 


Inspired by the above theory, we propose a hardware-friendly algorithm for 1-bit FQT. Our algorithm consists of an Activation Gradient Pruning (AGP) method to reduce gradient variance, which is an improvement based on per-group quantizers \cite{chen2020statistical, cho2020per}. AGP utilizes gradient heterogeneity by discarding less informative groups, and allocate saved resources to improve the numerical precision of more informative ones. 
Additionally, we propose Sample Channel joint Quantization (SCQ), an effective quantization scheme for accelerated performance. SCQ employs different quantization methods for computing weight gradients and activation gradients, ensuring both can be effectively implemented on low bitwidth computing units.



We examine the potential of 1-bit FQT on transfer learning tasks in both vision and NLP domain. In this task, 1-bit FQT algorithm is used for on-device finetuning a pretrained 1-bit model to adapt new data. On all the datasets, our 1-bit FQT algorithm can successfully \emph{converge} and demonstrate significantly superior performance compared to directly applying the previous FQT method to the task. The average accuracy drop on visual classification datasets is approximately 5\%, compared to training the binary model with full-precision gradients. Notably, the average accuracy loss is negligible (less than 1\%) on Flowers \cite{nilsback2008automated} dataset and Pets \cite{parkhi2012cats} dataset, indicating that 1-bit FQT might indeed be useful in some cases. 
We implement our algorithm on Hygon and Raspberry Pi devices as a PyTorch-based library \texttt{binop}. Accelerated on-device training can be achieved with simple layer substitution, e.g., replace \texttt{torch.nn.Conv2d} with  \texttt{binop.Conv2d}. In practice, our method can achieve up to 5.13× speedup, compared to FP32 PyTorch. These results indicate that in some specific tasks, FQT precision can be pushed to the ultimate 1-bit.




\section{Related Works}

\textbf{Quantization Aware Training.} 
QAT is a method designed to accelerate \emph{inference} by quantizing the activations and weights. 
Various works \cite{zhou2017incremental,choi2018pact,zhang2018lq,jacob2018quantization,dong2019hawq,tang2022mkq,liu2023llm}~
have been developed to quantize weights and activations into lower bitwidth. Furthermore, some studies \cite{rastegari2016xnor,bulat2019xnor,wang2020bidet,bai2020binarybert,wu2023estimator,qin2023distribution} have reduced the numerical precision of weights and activation values to 1 bit. However, QAT does not quantize gradients, and as a result, the backward propagation cannot be accelerated. 

\textbf{Fully Quantized Training.} FQT further quantizes the gradients into lower precision, compared with QAT. Hence, FQT allows for efficient implementation of both forward and backward propagation on low-bitwidth computational units. FQT, unlike optimizer quantization \cite{lin2022differentially}, involves quantizing weights, activations, and gradients altogether. Optimizer quantization only quantizes weight update (weight gradients), thus reducing communication costs but not accelerating computation \cite{saha2022efficient}. Early works on FQT use FP16 \cite{gupta2015deep,micikevicius2017mixed} or INT16 \cite{das2018mixed} values to constrain weights, activations, and gradients. After that, various 8-bit numerical formats \cite{wang2018training,banner2018scalable,zhu2020towards,yang2020training,xi2024jetfire} have been proposed that further push the bitwidth of data to 8 bits. Subsequently, \cite{chen2020statistical} provides theoretical bounds on how the quantization scheme (bitwidth, type of quantizer) affects the quality of the quantized gradient. Based on that, some works have successfully trained several networks with 4-bit activations/weights/gradients \cite{sun2020ultra,chmiel2021logarithmic,xi2023training}. The current research frontier is 4-bit FQT, but it still is not the ultimate limit.

\section{Framework}
To better describe our approach, necessary notations are introduced first. We denote the DNN model composed of $L$ layers with the learnable parameter $\boldsymbol{\Theta}$ as $\mathbf{F}(.;\boldsymbol{\Theta})$. In each training iteration, we sample a minibatch ($\mathbf{X},\mathbf{Y}$) from the dataset and input it into the model. The process is
\begin{equation}
\mathbf{H}^{(0)}=\mathbf{X}, \mathbf{H}^{(l)}=\mathbf{F}^{(l)}\left(\mathbf{H}^{(l-1)} ; \boldsymbol{\Theta}^{(l)}\right),  \forall l \in[L]_{+},
\label{eq1}
\end{equation}
where $\mathbf{H}^{(l)} \in \mathbb{R}^{N \times D^{(l)}}$ is a feature map ($N$ is the batch size, $D^{(l)}$ is the number of features), and $[L]_{+}=\{1,2, \ldots, L\}$ are sets of integers. $\mathbf{F}^{(l)}$ is the $l$-th layer of the model with parameter $\boldsymbol{\Theta}^{(l)}$.
Given the minibatch loss $\mathcal{L}(\mathbf{H}^{(L)}, \mathbf{Y})$, we compute the gradient $\nabla_{\boldsymbol{\Theta}^{(l)}} \mathcal{L}$, and update the parameter. For simplicity, we use $\nabla_{\mathbf{H}^{(l)}}$ and $\nabla_{\boldsymbol{\Theta}^{(l)}}$ represent the activation/parameter gradient. The back-propagation is
$\nabla_{\mathbf{H}^{(l-1)}}, \nabla_{\boldsymbol{\Theta}^{(l)}}=\mathbf{B}^{(l)}(\nabla_{\mathbf{H}^{(l)}}, \mathbf{H}^{(l-1)}, \boldsymbol{\Theta}^{(l)}),
$
where the function $\mathbf{B}^{(l)}$(·) takes the gradient of the output $\nabla_{\mathbf{H}^{(l)}}$ and the information kept in memory ($\mathbf{H}^{(l)}$, $\boldsymbol{\Theta}^{(l)}$), and computes the gradient of the input. For example, consider a linear layer $\mathbf{H}^{(l)}=\mathbf{H}^{(l-1)} \mathbf{\Theta}^{(l)}$ and its gradient is
\begin{equation}
\nabla_{\mathbf{H}^{(l-1)}}=\nabla_{\mathbf{H}^{(l)}} \boldsymbol{\Theta}^{(l)^{\top}}, \quad \nabla_{\boldsymbol{\Theta}^{(l)}}=\mathbf{H}^{(l-1)^{\top}} \nabla_{\mathbf{H}^{(l)}}.
\label{eq3}
\end{equation}

\subsection{Quantized Training}
Here, we describe Quantization-Aware Training (QAT) and Fully Quantized Training (FQT). QAT is employed to accelerate \emph{inference}, while FQT is designed to accelerate both inference and \emph{training}.\par
Before embarking on QAT, the initial step involves quantizing the parameters and activations of the model:
$\forall l \in[L]_{+},  \tilde{\mathbf{H}}^{(l-1)}=Q_f(\mathbf{H}^{(l-1)}),  \tilde{\boldsymbol{\Theta}}^{(l)}=Q_{\boldsymbol{\Theta}}(\boldsymbol{\Theta}^{(l)})
$,
where $Q_f$(·) and $Q_{\boldsymbol{\Theta}}$(·) are quantizers for activations and weights, and $\tilde{\mathbf{H}}^{(l-1)}$ and $\tilde{\boldsymbol{\Theta}}^{(l)}$ are quantized activations and weights. The forward propagation Eq. \ref{eq1} is quantized as
$\forall l \in[L]_{+}, \mathbf{H}^{(l)}=\mathbf{F}^{(l)}(\tilde{\mathbf{H}}^{(l-1)} ; \tilde{\boldsymbol{\Theta}}^{(l)}),
$
where $\tilde{\mathbf{H}}^{(l-1)}$ and $\tilde{\boldsymbol{\Theta}}^{(l)}$ represent low-bit data. Therefore, the inference can be efficiently implemented on low-bitwidth computing kernels. QAT leverages the straight-through estimator \cite{bengio2013estimating} to train quantized models.
The back-propagation Eq. \ref{eq3} becomes:
$
\tilde{\nabla}_{\mathbf{H}^{(l-1)}}=\nabla_{\mathbf{H}^{(l)}} \tilde{\boldsymbol{\Theta}}^{(l)^{\top}},  \tilde{\nabla}_{\boldsymbol{\Theta}^{(l)}}=\tilde{\mathbf{H}}^{(l-1)^{\top}} \nabla_{\mathbf{H}^{(l)}}.
$
Since gradients are not quantized, the backpropagation cannot be accelerated.\par
The forward propagation of FQT is identical to QAT, FQT further quantizes the gradients at each layer as
$
\forall l \in[L]_{+},   \hat{\nabla}_{\mathbf{H}^{(l)}}=Q_g\left({\nabla}_{\mathbf{H}^{(l)}}\right),$ where
$Q_g$(·) is a quantizer for gradients. The backpropagation is quantized as
$\hat{\nabla}_{\mathbf{H}^{(l-1)}}=\hat{\nabla}_{\mathbf{H}^{(l)}} \tilde{\boldsymbol{\Theta}}^{(l)^{\top}},  \hat{\nabla}_{\boldsymbol{\Theta}^{(l)}}=\tilde{\mathbf{H}}^{(l-1)^{\top}} \hat{\nabla}_{\mathbf{H}^{(l)}},$
where $\hat{\nabla}_{\mathbf{H}^{(l)}}$ and $\hat{\nabla}_{\boldsymbol{\Theta}^{(l)}}$ represent the FQT gradient. Now, with all operands quantized, the backpropagation can be efficiently implemented on low-bitwidth kernels.
\subsection{FQT with Unbiased Quantizer}\label{section3.2}
In our framework, $Q_f$(·) and $Q_{\boldsymbol{\Theta}}$(·) are deterministic quantizers, while $Q_g$(·) is an unbiased quantizer. This configuration follows \cite{chen2020statistical}. In this framework, the gradients in FQT are unbiased estimates of QAT, ensuring both converge to the same point in expectation. 
\par
Consider $Q_g$ as an unbiased stochastic quantizer, i.e., $\mathbb{E}\left[Q_g({\nabla}_{\mathbf{H}})\right]={\nabla}_{\mathbf{H}}$, for any ${\nabla}_{\mathbf{H}}$, which are already widely adopted in existing FQT approaches \cite{banner2018scalable,xi2023training}, thereby enabling $\mathbb{E}[\hat{\nabla}_{\mathbf{H}^{(l)}}]={\nabla}_{\mathbf{H}^{(l)}}$. The activation gradients of FQT is
$\mathbb{E}[\hat{\nabla}_{\mathbf{H}^{(l-1)}}]=\mathbb{E}[\hat{\nabla}_{\mathbf{H}^{(l)}}] \tilde{\boldsymbol{\Theta}}^{(l)^{\top}}=\tilde{\nabla}_{\mathbf{H}^{(l-1)}},$
which implies FQT and QAT convergence to a stationary point in expectation. Given an activation gradient tensor ${\nabla}_{\mathbf{H}}$, we quantize it to $b$-bit. We first compute the range of the tensor, and scale each element:
$\overline{\nabla}_{\mathbf{H}_{i,j}}=\operatorname{SR}(B({\nabla}_{\mathbf{H}_{i,j}}-Z)/R),$
where $B=2^b-1$ are the number of quantization bins, $R=\max \left\{{\nabla}_{\mathbf{H}}\right\}-\min \left\{{\nabla}_{\mathbf{H}}\right\}$ is the range, $Z=\min \left\{{\nabla}_{\mathbf{H}}\right\}$ is the zero point, the stochastic rounding \cite{courbariaux2015binaryconnect} operation $\operatorname{SR}$(·) convert input to integers, and $\overline{\nabla}_{\mathbf{H}_{i,j}}$ is the gradient quantized to $b$ bits. The dequantization is
$
\hat{\nabla}_{\mathbf{H}_{i,j}}=\overline{\nabla}_{\mathbf{H}_{i,j}} R/B+Z.
$
Due to the utilization of stochastic rounding, it is clear that $\mathbb{E}[\hat{\nabla}_{\mathbf{H}_{i,j}}]={\nabla}_{\mathbf{H}_{i,j}}$.\par

The unbiased quantizer widely adopted in FQT is the per-group quantizer, including per-tensor quantizer (PTQ) \cite{banner2018scalable}, per-sample quantizer (PSQ) \cite{chen2020statistical}, and per-channel quantizer (PCQ) \cite{cho2020per}. In these strategies, each group computes its own range and zero point, rather than sharing a common one, which addresses the large variation of dynamic range across groups.

\section{Theoretical Results} \label{theory}
In this section, we analyze the convergence behavior of FQT under two different optimizers, Adam and SGD. The proof of theorems follows the framework in \cite{kingma2014adam}, which can be found in Appendix \ref{appendix:proof}.
\subsection{Optimizer Impact on Convergence}
 Quantized training with the Adam optimizer achieved much higher accuracy than those with SGD (Fig. \ref{Adam-SGD}). Although some prior studies \cite{bulat2019xnor,lin2022device} have highlighted this issue, the theoretical understanding of FQT with Adam is still lacking. To fill this gap, we will provide theoretical bounds on the convergence of FQT based on both Adam and SGD optimizers in the following part.\par

 We use the framework proposed in \cite{zinkevich2003online} to analyze the convergence. We adopt the assumption made by \cite{zinkevich2003online} that the loss function $\mathcal{L}$ is convex. At each iteration $t$, we predict using the parameter ${\boldsymbol{\Theta}}_t$ and evaluate it on the loss function ${\mathcal{L}}_t$. We evaluate the convergence of FQT using the regret:
 $R(T)=\sum_{t=1}^T\left[{\mathcal{L}}_t\left({\boldsymbol{\Theta}}_t\right)-{\mathcal{L}}_t\left({\boldsymbol{\Theta}}^*\right)\right],$
where ${\boldsymbol{\Theta}}^*$ are the best fixed point parameter. We define ${\nabla}_{{\boldsymbol{\Theta}}_{1:t,i}} \in \mathbb{R}^t$ as a vector that contains the $i$-th dimension of the gradients over all iterations till $t$, ${\nabla}_{{\boldsymbol{\Theta}}_{1:t,i}}=[{\nabla}_{{\boldsymbol{\Theta}}_{1,i}},{\nabla}_{{\boldsymbol{\Theta}}_{2,i}},\ldots,{\nabla}_{{\boldsymbol{\Theta}}_{t,i}}]$,
$\hat{{\nabla}}_{{\boldsymbol{\Theta}}_{1:t,i}}$ is the quantized version of ${\nabla}_{{\boldsymbol{\Theta}}_{1:t,i}}$.
\begin{assumption}
There exists ${\sigma}, e > 0$, such that $\forall {\boldsymbol{\Theta}}_{t,i}$, $\operatorname{Var}\left[\hat{\nabla}_{\boldsymbol{\Theta}_{t,i}}\right] \leq \sigma^2$, $-e \leq \mathbb{E}\left[\hat{\nabla}_{\boldsymbol{\Theta}_{t,i}}\right] \leq e$.
\label{ass1}
\end{assumption}
\begin{assumption}
The distance between any $\boldsymbol{\Theta}_t$ is bounded, $\left\|\boldsymbol{\Theta}_n-\boldsymbol{\Theta}_m\right\|_2 \leq D$, $\left\|\boldsymbol{\Theta}_n-\boldsymbol{\Theta}_m\right\|_{\infty} \leq D_{\infty}, $ for any $m, n \in\{1, \ldots, T\}$.
\label{ass2}
\end{assumption}
Given an unbiased gradient, we now establish the convergence of quantized training under SGD. The iteration form of SGD is $\boldsymbol{\Theta}_{t+1} \leftarrow \boldsymbol{\Theta}_t-{\alpha}_t \hat{\nabla}_{{\boldsymbol{\Theta}}_t}$.
\begin{theorem}
\label{thm:bigtheorem}
If Assumption~\ref{ass1} and \ref{ass2} holds, let $\alpha_t=\frac{\alpha}{\sqrt{t}}$ and the number of elements in the gradient is $d$. SGD achieves the following guarantee, for all $T \geq 1$.
$R^{S G D}(T) \leq \frac{D^2}{2\alpha} + \frac{\alpha Td({\sigma}^2 + e^2)}{2}.$

\end{theorem}

The iteration form of Adam is expressed as follows:
$$
\begin{aligned}
    &\left\{
    \begin{array}{l}
        m_t=\beta_{1,t} \cdot m_{t-1}+\left(1-\beta_{1,t}\right) \cdot \hat{\nabla}_{{\boldsymbol{\Theta}}_t}, 
        v_t=\beta_2 \cdot v_{t-1}+\left(1-\beta_2\right) \cdot\left(\hat{\nabla}_{{\boldsymbol{\Theta}}_t}\right)^2, \\
        \hat{m}_t=\frac{m_t}{1-\beta_1^t}, \hat{v}_t=\frac{v_t}{1-\beta_2^t}, 
        \boldsymbol{\Theta}_{t+1}=\boldsymbol{\Theta}_t-\frac{\alpha}{\sqrt{\hat{v}}+\epsilon} \cdot \hat{m}_t.
    \end{array}
    \right.
\end{aligned}
$$
\begin{assumption}
The function ${\mathcal{L}}_t$ has bounded gradients, $\forall {\boldsymbol{\Theta}}$, $\left\|\hat{\nabla}_{{\boldsymbol{\Theta}}_t}\right\|_2 \leq G$, $\left\|\hat{\nabla}_{{\boldsymbol{\Theta}}_t}\right\|_{\infty} \leq G_{\infty}$.
\label{ass3}
\end{assumption}

\begin{theorem}
\label{thm:bigtheorem2}
If Assumption \ref{ass1}, \ref{ass2} and \ref{ass3} holds, let $\beta_1, \beta_2 \in[0,1)$ satisfy $\frac{\beta_1^2}{\sqrt{\beta_2}}<1$, $\alpha_t=\frac{\alpha}{\sqrt{t}}$, and $\beta_{1, t}=\beta_1 \lambda^{t-1}, \lambda \in(0,1)$. Adam achieves the following guarantee, for all $T \geq 1$.
$
\begin{aligned}
   R^{Adam}(T) \leq  \frac{((1-\lambda)^2D^2{T} + D_{\infty}^2)d}{2 \alpha\left(1-\beta_1\right)(1-\lambda)^2}  \sqrt{\sigma^2+e^2}
   +\frac{\alpha\left(1+\beta_1\right) G_{\infty}\sqrt{T}d}{\left(1-\beta_1\right) \sqrt{1-\beta_2}(1-\gamma)^2} \sqrt{\sigma^2+e^2}.
\end{aligned}
$
\end{theorem}
 Based on Theorem \ref{thm:bigtheorem}, \ref{thm:bigtheorem2}, Adam and SGD achieve the following guarantee, for $T \rightarrow \infty$.
$
\frac{R^{SGD}(T)}{T}\leq \alpha d(\sigma^2+e^2) / 2, 
\frac{R^{Adam}(T)}{T}\leq \frac{D^2d}{2 \alpha\left(1-\beta_1\right)}  \sqrt{\sigma^2+e^2}.$
From the inquation, it is straightforward to conclude that $\frac{R^{SGD}(T)}{T}=O(\sigma^2)+O(1)$, $\frac{R^{Adam}(T)}{T}=O(\sigma)+O(1)$. This implies that the convergence of FQT based on both Adam and SGD is influenced by the gradient variance, with SGD being more sensitive to variations in gradient variance.
\subsection{Quantizer Impact on Gradient Variance}

Based on our theory, gradient variance plays a crucial role in convergence. Gradient variance is primarily composed of two components: the variance of QAT gradients and the variance introduced by the gradient quantizers. \cite{chen2020statistical} reduced the complicated problem of gradient variance into the simple problem of quantizer variance. Thus, we need to minimize the quantizer variance.\par
The fundamental form of an unbiased quantizer $Q_g$ is given by Sec.  \ref{section3.2}, and its variance is
$
\begin{aligned}
\operatorname{Var}[Q_g(\hat{\nabla}_{\mathbf{H}^{(l)}}) \mid \hat{\nabla}_{\mathbf{H}^{(l)}}]=\frac{R^2}{B^2}\operatorname{Var}[\operatorname{SR}(\cdot) \mid \hat{\nabla}_{\mathbf{H}^{(l)}}]
\leq \frac{ND^{(l)}}{4 B^2} R^2,
\end{aligned}
$
where the maximum variance of stochastic rounding $\operatorname{SR}(\cdot)$ is 1/4. The expression reveals that as the bitwidth $b$ decreases, the variance significantly increases. Furthermore, due to the sensitivity of SGD to gradient variance, SGD performs less effectively than Adam in low precision scenarios (large gradient variance) (Fig. \ref{Adam-SGD}). Therefore, in scenarios with larger gradient variances, such as in quantized training, the Adam optimizer is recommended. Additionally, the variance is highly sensitive to the gradient range $R$, with outliers in the gradient expanding the range and consequently increasing the quantizer's variance.

\section{1-bit FQT Algorithm}
In this section, we propose our 1-bit FQT algorithm, including the quantization of weights, activation, and gradients.
\subsection{Forward Propagation}
In the forward propagation, both $Q_f$ and $Q_{\boldsymbol{\Theta}}$ are deterministic quantizers, taking the form:
$
\operatorname{sign}(x)=-1 \text { if } x \leq 0 \text { otherwise } 1 \text {. }
$
For a fully connected layer, the forward propagation is
$
{\mathbf{H}}^{(l)}=(\operatorname{sign}({\mathbf{H}}^{(l-1)})\operatorname{sign}({\boldsymbol{\Theta}}^{(l)}))\odot \Gamma,
$
where $\Gamma \in \mathbb{R}^{D^{(l)}}$ represents the shared scaling factor for both weights and activations, and it is learnable parameters. The form follows \cite{bulat2019xnor}.

\subsection{Backward Propagation} \label{backward}
The form of backpropagation is
\begin{equation}
        \hat{\nabla}_{\mathbf{H}^{(l-1)}}=Q_g(\hat{\nabla}_{\mathbf{H}^{(l)}}) \operatorname{sign}({\boldsymbol{\Theta}}^{(l)^{\top}}), 
        \hat{\nabla}_{\boldsymbol{\Theta}^{(l)}}=\operatorname{sign}({\mathbf{H}}^{(l-1)^{\top}}) Q_g(\hat{\nabla}_{\mathbf{H}^{(l)}}).
\label{eq21}
\end{equation}
Based on our theory, reducing quantizer variance is crucial to ensure the convergence of the model. However, outliers in the gradients can widen the range of gradients, thereby increasing variance.\par
To mitigate the impact of outliers on variance, per-group quantization is widely employed. Per-group quantization reduces variance by assigning a separate range to each group instead of sharing a large range among all. For example, we perform per-sample quantization on $\hat{\nabla}_{\mathbf{H}^{(l)}} \in \mathbb{R}^{N \times D^{(l)}}$ and its form is
$Q_g(\hat{\nabla}_{\mathbf{H}^{(l)}_{i,j}})=\operatorname{SR}(B(\hat{\nabla}_{\mathbf{H}^{(l)}_{i,j}}-Z_i)/R_i)R_i/B+Z_i,$
where $R_i, Z_i$ represent the range and zero point of activation gradients for the $i$-th sample.
Its variance is $
\operatorname{Var}[Q_g(\hat{\nabla}_{\mathbf{H}^{(l)}}) \mid \hat{\nabla}_{\mathbf{H}^{(l)}}]\leq \frac{D^{(l)}}{4 B^2} \sum_{i=1}^N R_{i}^2.$  However, the variance of PSQ is still too large for 1-bit FQT.\par
\begin{wrapfigure}[11]{r}{0.53\textwidth}
 \vskip -0.16in
  \centering
  \subfigure[]{
    \includegraphics[width=0.23\textwidth]{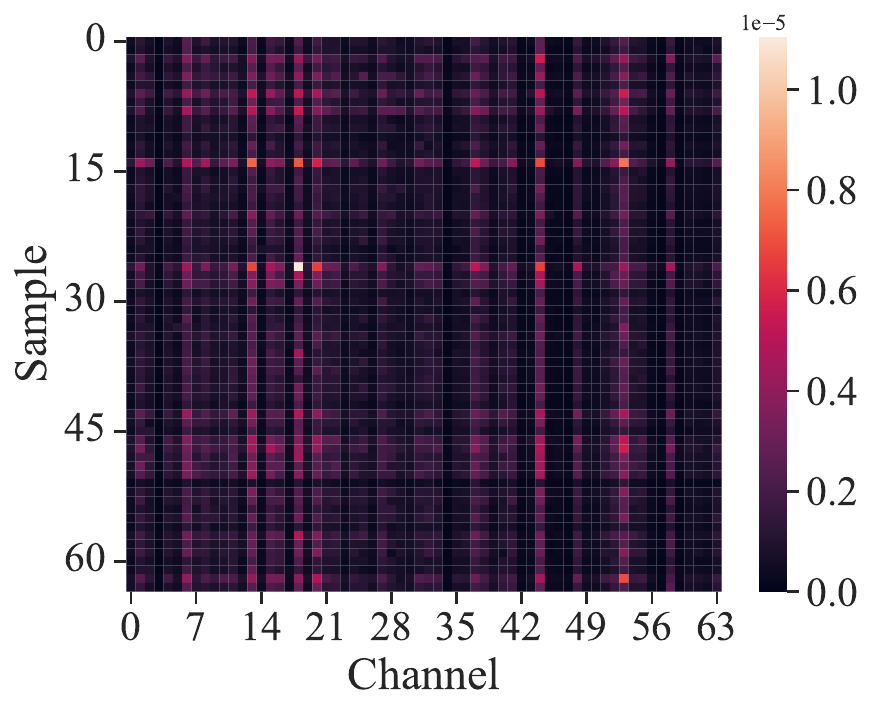}
    \label{dis:left}
  }
  \hspace{0.02\textwidth}  
  \subfigure[]{
    \includegraphics[width=0.23\textwidth]{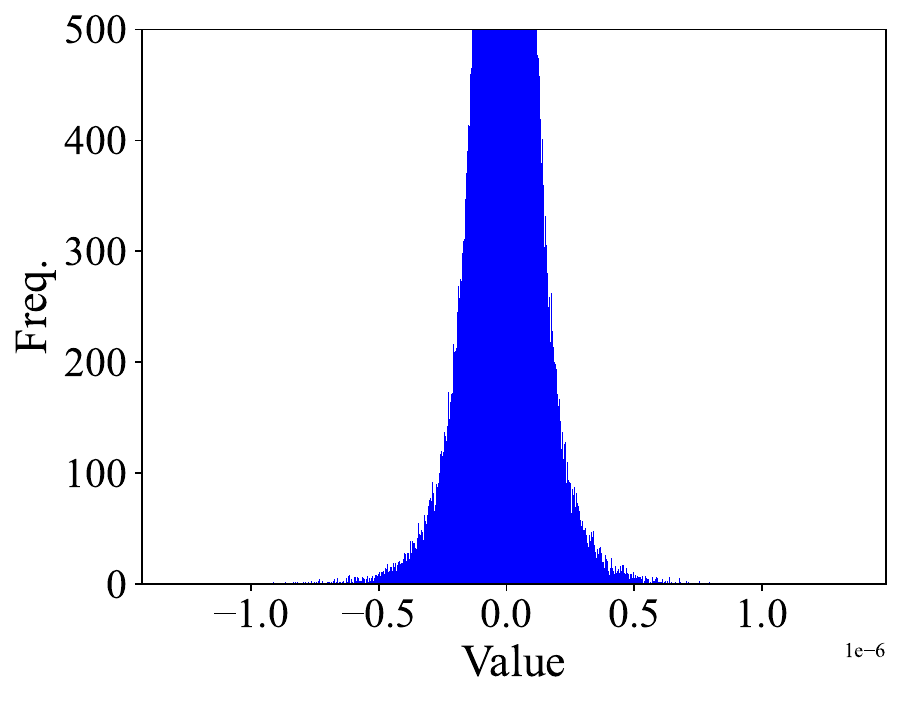}
    \label{dis:right}
  }
  \vspace{-10pt}
  \caption{Heterogeneity in a ResNet18’s gradients. (a) Heatmap
  of the per-group range at the conv2.1.2 layer; (b) Histogram of the
  gradient in a certain group.}
  \label{dis}
\end{wrapfigure}

 
To address this, we propose Activation Gradient Pruning (AGP) to reduce quantizer variance. The gradients exhibit heterogeneity \cite{xi2023training}, with some samples showing large gradient ranges, while the remaining gradients have smaller ranges. This pattern still holds true along the channel dimension, as illustrated in Fig. \ref{dis}. Additionally, from the visualizations of gradient distributions, we observe that the  elements in groups with smaller ranges are close to zero, indicating that less information stored in these groups. 
Therefore, we can prune groups with smaller ranges,
 reallocating the saved computational cost to groups with larger ranges, consequently enhancing the bitwidth of elements within those groups. Since variance mainly originates from groups with larger ranges, this approach effectively reduces variance.\par
Achieving 1-bit FQT based on the above idea requires ensuring three conditions: (1) if the bitwidth of retained groups is $b$, only $1/b$ of the groups can be preserved, thereby maintaining an average bitwidth of 1; (2) adopting random pruning to ensure the unbiased nature of quantization; (3) groups with larger ranges are more likely to be retained. Based on that, we first assign each group a probability $p_i \in[0,1], i=1, \cdots, N$. To retain $\frac{N}{b}$ groups and ensure the retained groups have a large range, $p_i$ needs to satisfy $\sum_{i=1}^{N} p_i=\frac{N}{b}$ and $p_i \propto R_i$, i.e., $p_i=\frac{NR_i}{bR_{total}}$, $R_{total}=\sum_{i=1}^{N}R_i$. Then we define random masks $m_i \sim \operatorname{Bern}\left(p_i\right)$ to prune unimportant groups, and perform per-group quantization on the remaining ones. Its form is:
$
Q_g(\hat{\nabla}_{\mathbf{H}^{(l)}})=Q_{PSQ}^b({\mathbf{M}}\hat{\nabla}_{\mathbf{H}^{(l)}}),
$
where ${\mathbf{M}}=\operatorname{diag}(\frac{m_1}{p_1}, \ldots, \frac{m_{N}}{p_{N}})$, $Q_{PSQ}^b$ is $b$-bit PSQ. $Q_g$ is an unbiased quantizer since $\mathbb{E}[Q_{PSQ}^b({\mathbf{M}}\hat{\nabla}_{\mathbf{H}^{(l)}})]=\mathbb{E}[{\mathbf{M}}] \hat{\nabla}_{\mathbf{H}^{(l)}}=\mathbf{I}\hat{\nabla}_{\mathbf{H}^{(l)}}$. The variance is
\begin{equation}
\operatorname{Var}\left[Q_g\left(\hat{\nabla}_{\mathbf{H}^{(l)}}\right) \mid \hat{\nabla}_{\mathbf{H}^{(l)}}\right]\leq \frac{D^{(l)}}{4 B^2} \sum_{i=1}^{\frac{N}{b}} R_{i}^2.
\label{eq24}
\end{equation}
From Eq. \ref{eq24}, it can be observed that the variance of our quantizer is significantly smaller than that of 1-bit PSQ ($\frac{D^{(l)}}{4} \sum_{i=1}^N R_{i}^2$). The proof is given
in Appendix \ref{appendix:variance}.\par
\begin{figure*}[t]
\vskip -0.2in
\begin{center}
\centerline{\includegraphics[width=\textwidth]{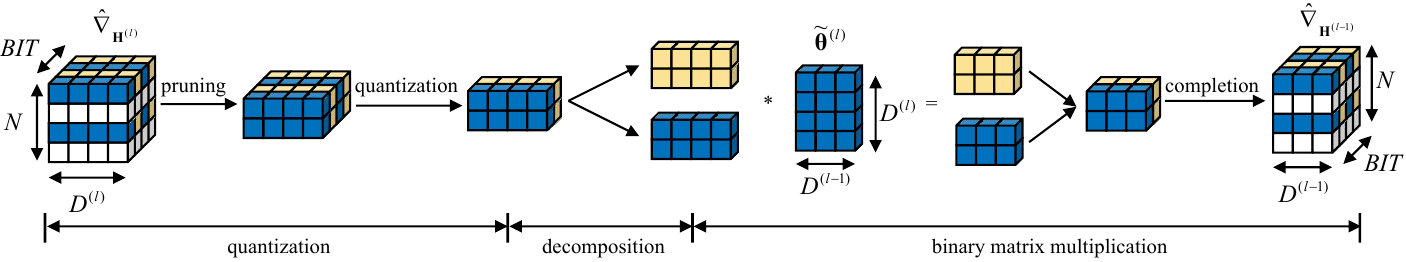}}
\caption{The process of AGP and binary matrix multiplication. $BIT$ represents the bitwidth of full precision data. Here, we removed half of the groups, thus the bitwidth of the remaining groups is 2.}
\label{agp}
\end{center}
\vskip -0.1in
\end{figure*}

Despite reducing the average precision of each element to 1 bit through pruning, the practical acceleration is hindered by the fact that the retained groups maintain a precision of $b$-bit. We perform a lossless decomposition of the $b$-bit element: $x=\sum_{i=1}^{b}2^{(i-1)}x^i$, where $x^i$ represents the value at $i$-th bit of the $b$-bit element. Extend the decomposition operation to the entire tensor, Eq. \ref{eq21} becomes:$\hat{\nabla}_{\mathbf{H}^{(l-1)}}=\sum_{i=1}^b{2^{(i-1)}{(Q_g(\hat{\nabla}_{\mathbf{H}^{(l)}}))}^i \operatorname{sign}({\boldsymbol{\Theta}}^{(l)^{\top}})},\hat{\nabla}_{\boldsymbol{\Theta}^{(l)}}=\sum_{i=1}^b{2^{(i-1)}{\operatorname{sign}({\mathbf{H}}^{(l-1)^{\top}}) {(Q_g(\hat{\nabla}_{\mathbf{H}^{(l)}}))}^i}},$ where ${(Q_g(\hat{\nabla}_{\mathbf{H}^{(l)}}))}^i$ represents the $i$-th binarized slice tensor of ${Q_g(\hat{\nabla}_{\mathbf{H}^{(l)}})}$, where each element represents the value at the $i$-th bit of the corresponding element in ${Q_g(\hat{\nabla}_{\mathbf{H}^{(l)}})}$. Due to the removal of some groups, the shape of the result differs from the original, and we fill the gaps with zeros. The entire process is illustrated in Fig. \ref{agp}.

\subsection{Practical Acceleration}
In order to ensure the compatibility of our method with low-bit hardware, we first illustrate the necessary conditions for achieving practical acceleration through a PSQ example. 
The form of PSQ can be rewritten as: $Q_{PSQ}(\hat{\nabla}_{\mathbf{H}^{(l)}})=\mathbf{S}_{float}\overline{\nabla}_{\mathbf{H}^{(l)}}$, where $\mathbf{S}_{float}=\operatorname{diag}\left\{\frac{R_1}{B}, \ldots, \frac{R_N}{B}\right\}$ is a FP32 tensor and $\overline{\nabla}_{\mathbf{H}^{(l)}}$ is a 1-bit tensor. The Eq. \ref{eq21} can be rewritten as: $\hat{\nabla}_{\mathbf{H}^{(l-1)}}=\mathbf{S}_{float}\overline{\nabla}_{\mathbf{H}^{(l)}} \operatorname{sign}({\boldsymbol{\Theta}}^{(l)^{\top}}),
        \hat{\nabla}_{\boldsymbol{\Theta}^{(l)}}=\operatorname{sign}({\mathbf{H}}^{(l-1)^{\top}}) \mathbf{S}_{float}\overline{\nabla}_{\mathbf{H}^{(l)}},$
\begin{figure*}[t]
\vskip -0.1in
\begin{center}
\centerline{\includegraphics[width=\textwidth]{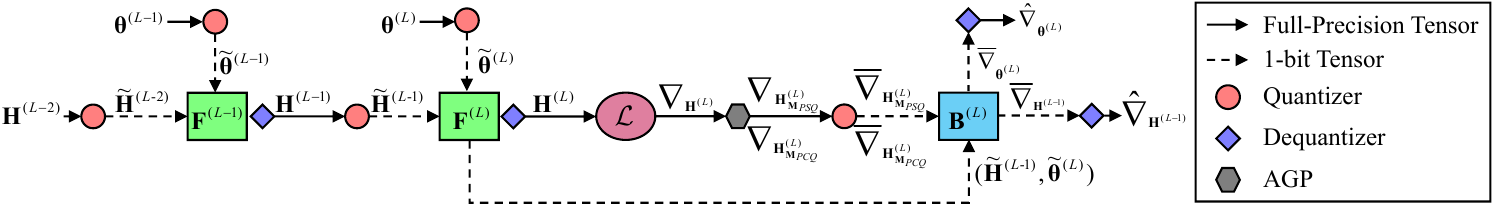}}
\caption{The computational graph of 1-bit FQT. The figure only shows the forward propagation of the last two layers and the backward propagation of the final layer.}
\label{process}
\end{center}
\vskip -0.33in
\end{figure*}
where the computation of activation gradients $\hat{\nabla}_{\mathbf{H}^{(l-1)}}$ can be accelerated, as 1-bit matrix multiplication (MM) $\overline{\nabla}_{\mathbf{H}^{(l)}}\operatorname{sign}({\boldsymbol{\Theta}}^{(l)^{\top}})$ can be implemented efficiently on hardware. However, weight gradients $\hat{\nabla}_{\boldsymbol{\Theta}^{(l)}}$ cannot be efficiently computed, as one of the two 1-bit tensors $\overline{\nabla}_{\mathbf{H}^{(l)}}$ needs to be dequantized to a full-precision tensor before their multiplication. This severely limits the acceleration.\par
To address this issue, we propose Sample Channel joint Quantization (SCQ), wherein PCQ is employed during the computation of weight gradients, while PSQ is utilized for the computation of activation gradients. Building upon this quantization strategy, the backpropagation process can be rewritten as:
$\hat{\nabla}_{\mathbf{H}^{(l-1)}}=\mathbf{S}_{PSQ}^{}\overline{\nabla}_{\mathbf{H}_{PSQ}^{(l)}} \operatorname{sign}({\boldsymbol{\Theta}}^{(l)^{\top}}), \hat{\nabla}_{\boldsymbol{\Theta}^{(l)}}=\operatorname{sign}({\mathbf{H}}^{(l-1)^{\top}}) \overline{\nabla}_{\mathbf{H}_{PCQ}^{(l)}}\mathbf{S}_{PCQ}^{},$
where $\mathbf{S}_{PCQ}^{}=\operatorname{diag}\left\{\frac{R_1^c}{B}, \ldots, \frac{R^c_{D^{(l)}}}{B}\right\}$, $R_i^c$ represents the range of $i$-th channel. This strategy facilitates the acceleration of both weight and activation gradient computations. We integrate this strategy with the previously proposed AGP (Fig. \ref{process}), resulting in the final formulation:
$\hat{\nabla}_{\mathbf{H}^{(l-1)}}=Q_{PSQ}^b\left({\mathbf{M}}_{PSQ}\hat{\nabla}_{\mathbf{H}^{(l)}}\right) \operatorname{sign}({\boldsymbol{\Theta}}^{(l)^{\top}}),
        \hat{\nabla}_{\boldsymbol{\Theta}^{(l)}}=\operatorname{sign}({\mathbf{H}}^{(l-1)^{\top}}) Q_{PCQ}^b\left(\hat{\nabla}_{\mathbf{H}^{(l)}}{\mathbf{M}}_{PCQ}\right).$
Since PCQ treats a channel as a group, pruning operations also need to be performed along the channel dimension. Due to space constraints, decomposition and implementation details are provided in Appendix \ref{appendix:implementation}. 

\section{Experiments}\label{Experiment}
We evaluate our approach on transfer learning tasks. Although our approach is constrained to transfer learning, it still holds practical value in on-device training \cite{lin2022device}. Due to challenges such as environmental constraints and limited memory, it is impractical to perform training from scratch on edge devices \cite{ren2021tinyol}.  The experiment details and results from training from scratch are in Appendix \ref{appendix:experiment}.
\begin{table*}[t]
\vskip -0.2in
\caption{Experimental results on multiple downstream datasets. ``(W, A, G)" denote the bitwidth of weight, activations, and gradients, respectively. $b$ represents the bitwidth of the remaining groups.}

\small
	\centering

    \setlength{\tabcolsep}{0.8mm}
	\begin{tabular}{l c c c c c c c r}
		\toprule
		
		\multirow{2}{*}{Method} & {Precision} & \multicolumn{7}{c}{Accuracy(\%)} \\
        \cline{3-9}
        &{(W, A, G)}& CIFAR-10 & CIFAR-100 & Flowers & Cars & Pets & CUB & Average   \\
		
		\midrule
		\multicolumn{9}{c}{ResNet-18}\\
		QAT &1, 1, 32 & 87.31$\pm$.25  & 65.82$\pm$.43 & 78.85$\pm$.80 & 50.81$\pm$.38 & 71.68$\pm$.21 & 42.13$\pm$.43 & 66.10\\
  		PSQ &8, 8, 8 & 92.90$\pm$.03 & 75.83$\pm$.05 & 85.96$\pm$.44 & 71.09$\pm$.71 & 87.86$\pm$.20 & 61.22$\pm$.05 & 79.14\\
        \hdashline
        PSQ &1, 1, 1 &71.04$\pm$.61&	47.71$\pm$.98&	78.91$\pm$.10&		23.14$\pm$.91&	68.93$\pm$.39&	34.29$\pm$.62& 54.01\\
        Ours ($b=2$) &1, 1, 1 & 74.10$\pm$.21&	52.19$\pm$.62&	\textbf{79.93}$\pm$.20&	26.51$\pm$.76&	70.47$\pm$.52&	36.59$\pm$.31&	56.63\\
        Ours ($b=4$) &1, 1, 1 & \textbf{78.52}$\pm$.56&	\textbf{56.83}$\pm$.61&	79.28$\pm$.50&	\textbf{37.88}$\pm$.36&	\textbf{71.17}$\pm$.16&	\textbf{39.47}$\pm$.25&	\textbf{60.53}\\
        Ours ($b=8$) &1, 1, 1 &73.73$\pm$.99&	52.64$\pm$.36&	78.10$\pm$.65&	29.78$\pm$.89&	69.98$\pm$.32&	37.01$\pm$.53&	56.87\\

        \midrule
		\multicolumn{9}{c}{VGGNet-16}\\
        QAT &1, 1, 32 & 89.80$\pm$.36 & 71.70$\pm$.17 & 86.86$\pm$.35 & 67.65$\pm$.03 & 79.49$\pm$.44 & 53.39$\pm$.57 & 74.82\\
        PSQ & 8, 8, 8 & 90.88$\pm$.07	&73.12$\pm$.09&	88.59$\pm$.15	&	81.27$\pm$.05	&90.33$\pm$.03	&69.49$\pm$.09&	82.28\\
        \hdashline
        PSQ &1, 1, 1 &80.60$\pm$.20&	59.81$\pm$.20&	84.65$\pm$.05&	40.01$\pm$.88&	77.20$\pm$.38&	43.17$\pm$.44&	64.24\\
        Ours ($b=2$) &1, 1, 1 & 82.66$\pm$.44&	62.04$\pm$.01&	85.75$\pm$.29&	44.40$\pm$.92&	77.77$\pm$.35&	46.33$\pm$.53&	66.49\\
        Ours ($b=4$) &1, 1, 1 & \textbf{84.38}$\pm$.12&	\textbf{63.65}$\pm$.19&	\textbf{87.12}$\pm$.20&	\textbf{57.06}$\pm$.60&	\textbf{78.48}$\pm$.21&	\textbf{49.10}$\pm$.17&	\textbf{69.97}\\
        Ours ($b=8$) &1, 1, 1 &78.14$\pm$.86&	60.20$\pm$.08&	86.24$\pm$.15&	46.95$\pm$.21&	77.39$\pm$.26&	47.48$\pm$.20&	66.07\\
		\bottomrule
	\end{tabular}
 \vskip -0.05in
\label{table1}
\end{table*}

\begin{table*}[t]
\caption{Comparison of training speedup across different input resolutions. "-Basic" and "Basic" represent unoptimized FQT and unoptimized FP32 training. The baseline is FP32 Pytorch.}
\small
    \centering
    \setlength{\tabcolsep}{1.5mm}
    \begin{tabular}{l c c c  c c   c c| c  c c c}
        \toprule
        
        \multirow{2}{*}{Model} & \multirow{2}{*}{Method} &{Precision}& \multicolumn{5}{c}{Hygon} & \multicolumn{3}{c}{Raspberry Pi 5}\\
        \cline{4-11}
        &&(W, A, G)& 32 & 64 & 128 & 224 & average& 32 & 64 & average\\
        \midrule
        \multirow{4}{*}{VGGNet-16} & Basic & 32, 32, 32 &0.04× & 0.03× & 0.03× & 0.02× & 0.03× & 0.05× & 0.02× & 0.04×\\
        & PSQ-Basic & 8, 8, 8&0.08× & 0.06× & 0.06× & 0.06× & 0.07× & 0.10× & 0.04× & 0.07×\\
        & SCQ-Basic & 8, 8, 8& 0.24× & 0.18× & 0.17× & 0.15× & 0.19× & 0.23× & 0.08× & 0.16×\\
        & Ours-Basic & 1, 1, 1& 4.36× & 3.25× & 3.19× & 1.86× & 3.17× & 3.47× & 1.17× & 2.32×\\
        & Ours & 1, 1, 1& 5.13× & 3.71× & 3.38× & 2.73× & 3.74× & 3.72× & 1.25× & 2.49×\\

        \midrule
        \multirow{4}{*}{ResNet-18} & Basic & 32, 32, 32& 0.03× & 0.03× & 0.03× & 0.02× & 0.03× & 0.02× & 0.02× & 0.02×\\  
        & PSQ-Basic & 8, 8, 8& 0.07× & 0.07× & 0.06× & 0.06× & 0.07× & 0.05× & 0.03× & 0.04×\\   
        & SCQ-Basic & 8, 8, 8& 0.17× & 0.17× & 0.16× & 0.16× & 0.17× & 0.11× & 0.08× & 0.10×\\ 
        & Ours-Basic & 1, 1, 1& 2.69× & 2.75× & 2.31× & 1.30× & 2.26× & 1.33× & 0.92× & 1.13×\\
        & Ours &1, 1, 1& 2.93× & 2.88× & 2.62× & 2.15× & 2.65× & 1.42× & 0.97× & 1.20×\\   
        \bottomrule
    \end{tabular}
 \vskip -0.15in
\label{table3}
\end{table*}
\subsection{Main Results} \label{main_result}
We employed two DNN architectures, ResNet18 \cite{he2016deep} and VGGNet16 \cite{simonyan2014very}. We pre-trained them on ImageNet \cite{deng2009imagenet} and subsequently conducted QAT. The quantized models are fine-tuned on downstream datasets to evaluate our approach. Following \cite{lin2022device}, we utilize various datasets, including Cars \cite{krause20133d}, CIFAR-10 \cite{krizhevsky2009learning}, CIFAR-100 \cite{krizhevsky2009learning}, CUB \cite{welinder2010caltech}, Flowers \cite{nilsback2008automated} and Pets \cite{parkhi2012cats}. \par

\textbf{Converged model accuracy.}
To evaluate the performance of our method, we report the accuracy of two model architectures, VGG16 and ResNet18, across various datasets in Table \ref{table1}. We report the mean and stddev of 3 runs. The compared approaches include QAT \cite{bulat2019xnor} and PSQ. Since QAT employs training with full precision gradients, it can be considered as an upper bound for the accuracy of 1-bit FQT. Existing work has not tried 1-bit FQT, so we did not compare more methods. On VGGNet16, our method achieves $< 10\%$ average accuracy degradation across all configurations, as compared to the baseline QAT with 32-bit gradients. Moreover, in the optimal configuration ($b$=4), our method exhibit only approximately 5\% average accuracy drop. On the more challenging ResNet18, the worst configuration ($b$=2) and the optimal configuration ($b$=4) achieves 9.47\% and 5.57\% average accuracy degradation, respectively, compared to QAT. Furthermore, on some datasets such as Flowers and Pets, our method exhibits minimal accuracy loss, indicating its suitability for these datasets. In summary, while our approach exhibits a notable decrease in accuracy compared to QAT, the incurred gap remains acceptable considering the benefits gained from reducing the numerical precision of gradients to 1 bit. Additionally, we compared our method with 1-bit PSQ. Across both frameworks, our approach consistently outperformed it in terms of average accuracy across all configurations. Moreover, except for individual outcomes in the worst configuration, our method also exhibited superior accuracy across all datasets.\par

\begin{figure}[t]
\vskip -0.1in
\centering
\begin{minipage}[b]{0.48\textwidth}
  \centering
  \includegraphics[width=\textwidth]{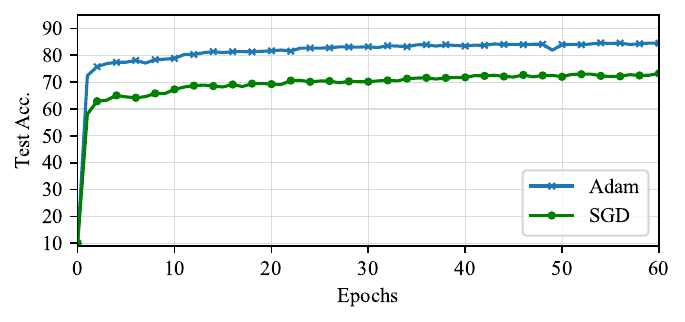}
  \label{fig3:left}
  
\end{minipage}
\hfill
\begin{minipage}[b]{0.48\textwidth}
  \centering
  \includegraphics[width=\textwidth]{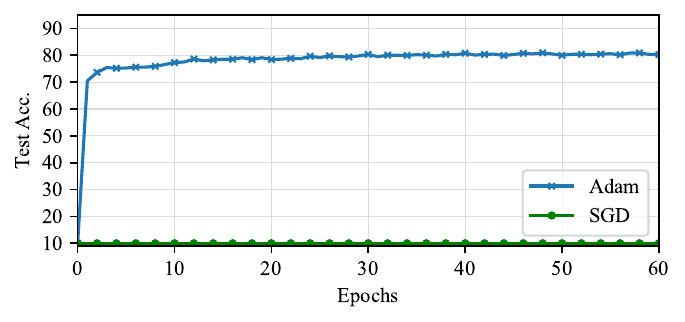}
  \label{fig3:right}
  
\end{minipage}
\caption{Our method (left) vs. PSQ (right): Testing accuracy on VGGNet16 for CIFAR-10.}
\vskip -0.2in
\label{fig3}
\end{figure}
\textbf{The value of $b$.}
We investigate the impact of hyperparameter $b$ on performance and determine the optimal choice for $b$. From Eq. \ref{eq24}, as $b$ increases, the variance of the quantizer gradually decreases, suggesting an improvement in training convergence. However, the increase in $b$ also implies more discarded groups, leading to larger losses. Therefore, the choice of $b$ becomes a trade-off issue. In Table \ref{table1}, we report the accuracy of our method across various datasets under three different configurations ($b=2$, $b=4$, and $b=8$). On VGGNet16 and ResNet18, the configuration with $b=4$ consistently outperforms the others ($b=2$ and $b=8$) in terms of average accuracy. Moreover, this observation extends to the majority of datasets, where, even on a few datasets, the results for the configuration
\begin{wrapfigure}[13]{r}{0.35\textwidth}
\vskip -0.20in
  \centering
  \includegraphics[width=0.35\textwidth]{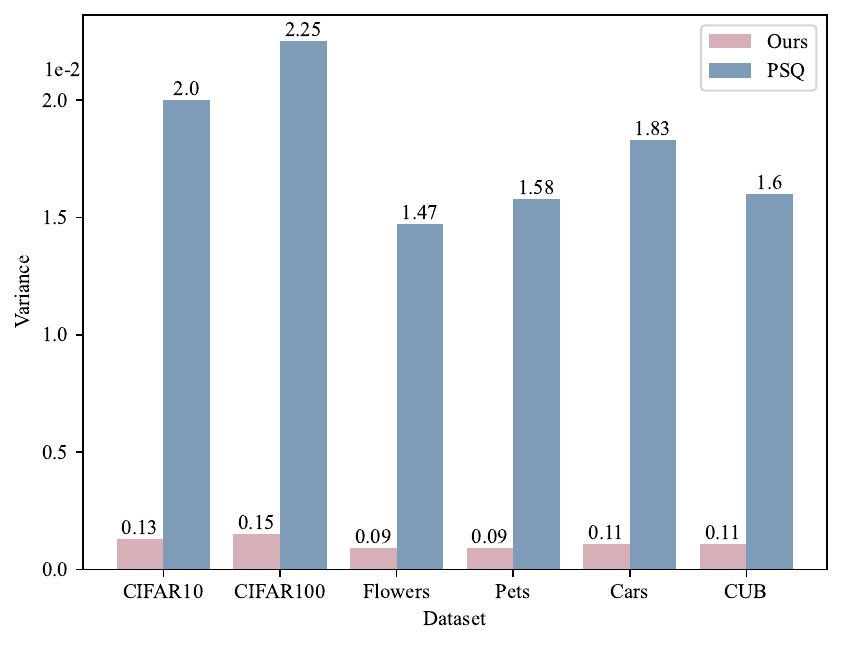}
  \vspace{-10pt}
  \caption{Quantizer variances across different datasets.}
  \label{fig11}
  \vskip -0.2in
\end{wrapfigure}
with $b=4$ may not be optimal, the performance difference remains marginal compared to the optimal results. In conclusion, the optimal configuration is $b=4$.\par

\textbf{Effect of the optimizer.}
To validate our theory that the SGD optimizer is more sensitive to the variance of gradients compared to the Adam optimizer, we conduct a performance comparison of different optimizers on the CIFAR-10 dataset. We present 
the test accuracy curves of our method and PSQ across
different optimizers in Fig. \ref{fig3}. For both methods, model performance degrades when using the SGD compared to the Adam. This is primarily attributed to the sensitivity of SGD to gradient variance. In addition, we observed that our method with SGD experienced only a modest accuracy drop, whereas the PSQ method with SGD failed to converge entirely. We attribute this observation to the larger variance introduced by PSQ compared to our quantizer, resulting in divergence.\par

\textbf{Variance.}To demonstrate the advantages of our quantizer in reducing variance, we present the quantizer variance of ResNet18 in 
Fig. \ref{fig11}. In general, the quantizer variance of our method is lower than that of PSQ across all datasets. Additionally, the variance on the Flowers and Pets is lowest, explaining why the impact of quantization on accuracy is minor for them.\par

\textbf{Other results.}
We report results for other architectures and tasks in Table \ref{table2}. The details can be found in Appendix \ref{appendix:experiment}. On Faster R-CNN \cite{ren2015faster}, our approach with 1-bit gradients achieves 1.66\% mAP degradation, as compared to the baseline QAT with 32-bit gradients. In addition, for Mixer-MLP \cite{tolstikhin2021mlp}, an all-MLP architecture, our approach shows a decrease of 3.52\% in classification accuracy compared to the baseline. For BERT, our approach achieves 8.39\% average performance degradation. These results indicate the potential of our approach to transfer to other architectures and tasks.\par

\subsection{Computational Efficiency}
We discuss the computational overhead of our method. Our implementation is not fully optimized, as the comprehensive hardware-algorithm co-design is beyond the scope of this paper. Our experiments are conducted on a single-core Hygon CPU and edge device (Raspberry Pi 5). 
\par
\textbf{Training speedup.} We compare the training time of the FP32 PyTorch and our 1-bit FQT for VGGNet16 and ResNet18. We vary the resolution of the input and summarize the speedup of our method in Table \ref{table3}. For VGGNet16, our algorithm achieves an average speedup of 3.74× and 2.49× on the Hygon and edge device, respectively. For ResNet18, our algorithm achieves 2.65× and 1.20× average speedup. Additionally, to assess the acceleration potential of 1-bit FQT, we compare their speedup under the condition without optimization (Ours-Basic and Basic). The results indicate that across multiple cases, the speedup is well above a \emph{hundredfold}. On edge devices, our method achieves a speedup of over 50×. This gap indicates significant acceleration potential for 1-bit FQT. Finally, we compare the speedup across layers of VGGNet16 and the time consumption for each operation in Appendix \ref{appendix:experiment}, providing guidance for future optimization directions.\par
\textbf{8-bit PSQ vs. Ours.} 
To demonstrate the advantages of our method over other high-bit-width FQT methods, we compare our approach with 8-bit PSQ in terms of both speedup and classification performance (there is no 4-bit format among the standard data types). Table \ref{table1} reveals that compared to 8-bit PSQ, our method experiences an average accuracy reduction of 18.61\% and 12.31\% on ResNet18 and VGG16, respectively. Despite significant performance degradation, our approach achieves an average speedup of 32.28× and 45.28× over 8-bit PSQ on ResNet18 and VGG16, respectively. On Raspberry Pi 5, our speedup can also reach approximately 30×. Even when compared to hardware-friendly SCQ, 1-bit FQT achieves a speedup of over 10× (Table \ref{table3}) Therefore, our method can be particularly beneficial in scenarios sensitive to time overhead.
\par
\textbf{Average 1-bit vs. 1-bit.} 
We compared the runtime of average 1-bit matrix multiplication and 1-bit matrix multiplication across different matrix sizes in Table \ref{table5}. The results demonstrate that the difference in runtime between these two methods is minimal, indicating similarity in the runtime of our average 1-bit FQT and 1-bit FQT. The analysis is provided in the Appendix \ref{appendix:implementation}.
\begin{table*}[t]
\vspace{-10pt}
\centering
\begin{minipage}[c]{0.50\textwidth}
\small
\caption{Object detection on PASCAL VOC, classification on CIFAR-100 and NLP tasks on GLUE.}
\label{table2}
\setlength{\tabcolsep}{0.65mm}
\centering
\begin{tabular}{c c c c c}
\toprule

{Task}& {Model}& {Method} & {Bits} &{mAP/Acc./Avg.} \\
\midrule
\multirow{2}{*}{Det.}& Faster R-CNN & QAT & 32 & 52.34\\
& Faster R-CNN & Ours & 1 & 50.68\\
\midrule
\multirow{2}{*}{Cls.}& MLP-Mixer & QAT & 32 & 52.17\\
& MLP-Mixer & Ours & 1 & 48.65\\	
\midrule
\multirow{2}{*}{NLP}& BERT & QAT & 32 & 63.20\\
& BERT & Ours & 1 & 54.81\\        
\bottomrule
\end{tabular}

\end{minipage}
\hfill
\begin{minipage}[c]{0.47\textwidth}
\caption{Average 1-bit vs. 1-bit. The running time of matrix multiplication involving various sizes.}
\label{table5}
\small
\centering
\setlength{\tabcolsep}{0.5mm}
\begin{tabular}{c c c c c c c}
    \toprule
    
    \multirow{4}{*}{Setting} &\multicolumn{6}{c}{Time(ms) across various sizes}\\
    \cline{2-7}
    & 512& 512 &1024&1024&2048&2048\\
    & 512& 512 &512&512&512&512\\
    & 512& 1024 & 512& 1024 & 512& 1024 \\
    
    \midrule
    Average 1-bit& 8.40	&15.92		&16.61&	31.03	&32.22	&61.33	\\
    1-bit & 8.01& 14.54&	15.91	&	29.79	&29.92	&59.94\\		
    \bottomrule
\end{tabular}
\vskip -0.1in
\end{minipage}
\vspace{-10pt}
\end{table*}

\section{Conclusion} \label{conclusion}
We propose a hardware-friendly 1-bit FQT method in this work, which pushes the limit of FQT. Through convergence analysis, we propose AGP to reduce the variance of the quantizer, thereby enhancing the convergence of quantized training. Subsequently, to address the issue of unacceleratable weight gradient computation, we present a SCQ strategy. Finally, we propose a framework that practically accelerates training, achieving a speedup of up to 5.13× compared to full precision training. While our approach focuses solely on convolutional neural networks in this study, experiments indicate its potential applicability to other architectures.\par
\textbf{Limitations:} The primary limitation of this work lies in its ability to achieve 1-bit FQT in transfer learning tasks but not in training from scratch. To the best of our knowledge, even the 3-bit FQT from scratch is still an open problem.

\section* {Broader Impact}
The introduction of 1-bit Fully Quantized Training (FQT) presents a significant advancement in deep neural network training, promising accelerated model training. Recently, huge models \cite{brown2020language} have achieved remarkable results across various domains. However, the training cost of these models is becoming increasingly expensive. The escalating training costs are leading to a growing dependence on computational resources in machine learning research, which raises concerns about fairness. FQT can effectively reduce the training expenses, thereby democratizing model training for a broader audience. Additionally, 1-bit FQT makes training on edge devices feasible. The models deployed on edge devices require updates when learning new data. However, due to environmental and memory constraints, these models cannot be directly updated \cite{ren2021tinyol}. One straightforward approach is to conduct training directly on the edge device. However, the computational resources of edge devices are often insufficient to support model training. 1-bit FQT holds the promise of addressing this challenge.

\medskip

{
\small

\bibliographystyle{plainnat}
\bibliography{example_paper}
}

\newpage
\appendix

\section{Proof of Theorems}\label{appendix:proof}
\begin{lemma}
\label{lem:1}
If a function $\mathcal{L}: R^d \rightarrow R$ is convex, then for all $x, y \in R^d$,
$$
\mathcal{L}(y) \geq \mathcal{L}(x)+\nabla \mathcal{L}(x)^T(y-x).
$$
\end{lemma}

\begin{lemma}
\label{lem:2}
Let $\hat{\nabla}_{{\boldsymbol{\Theta}}_t}=\hat{\nabla} \mathcal{L}_t\left(\boldsymbol{\Theta}_t\right)$ and $\hat{\nabla}_{{\boldsymbol{\Theta}}_{1: t}}$ be defined as above and bounded, $\left\|\hat{\nabla}_{{\boldsymbol{\Theta}}_t}\right\|_2 \leq G,\left\|\hat{\nabla}_{{\boldsymbol{\Theta}}_t}\right\|_{\infty} \leq$ $G_{\infty}$. Then,
$$
\sum_{t=1}^T \sqrt{\frac{\hat{\nabla}_{{\boldsymbol{\Theta}}_{t, i}^2}}{t}} \leq 2 G_{\infty}\left\|\hat{\nabla}_{{\boldsymbol{\Theta}}_{1: T, i}}\right\|_2.
$$
\end{lemma}

\begin{lemma}
\label{lem:3}
Let $\gamma \triangleq \frac{\beta_1^2}{\sqrt{\beta_2}}$. For $\beta_1, \beta_2 \in[0,1)$ that satisfy $\frac{\beta_1^2}{\sqrt{\beta_2}}<1$ and bounded $\hat{\nabla}_{{\boldsymbol{\Theta}}_{t}},\left\|\hat{\nabla}_{{\boldsymbol{\Theta}}_{t}}\right\|_2 \leq G$, $\left\|\hat{\nabla}_{{\boldsymbol{\Theta}}_{t}}\right\|_{\infty} \leq G_{\infty}$, the following inequality holds
$$
\sum_{t=1}^T \frac{\widehat{m}_{t, i}^2}{\sqrt{t \widehat{v}_{t, i}}} \leq \frac{2}{1-\gamma} \frac{1}{\sqrt{1-\beta_2}}\left\|\hat{\nabla}_{{\boldsymbol{\Theta}}_{1: T, i}}\right\|_2.
$$
\end{lemma}
The above lemma has been previously proven in \cite{kingma2014adam}, and we omit its reproof here for brevity.
\begin{lemma}
\label{lem:4}
For a random matrix $\mathbf{X}$, the following inequality holds
$$
\mathbb{E}[\|\mathbf{X}\|_2] \leq \sqrt{\mathbb{E}[\|\mathbf{X}\|_2^2]}
$$
\end{lemma}
\textit{Proof.} According to the formula $\mathbb{E}[{x}^2]=\operatorname{Var}[x] + {\mathbb{E}^2[x]}$, we can derive:
$$
\begin{aligned}
\sqrt{\mathbb{E}[\|\mathbf{X}\|_2^2]} & = \sqrt{\mathbb{E}^2[\|\mathbf{X}\|_2] + \operatorname{Var}[\|\mathbf{X}\|_2]}\\
& \geq \sqrt{\mathbb{E}^2[\|\mathbf{X}\|_2]}\\
& = \mathbb{E}[\|\mathbf{X}\|_2].
\end{aligned}
$$
\subsection{Theorem 4.3: Convergence of SGD}
\textit{Proof.} The iteration form of SGD is $$\boldsymbol{\Theta}_{t+1} \leftarrow \boldsymbol{\Theta}_t-{\alpha}_t \hat{\nabla}_{{\boldsymbol{\Theta}}_t}.$$
Subtract the scalar $\boldsymbol{\Theta}^*$ and square both sides of the above update, we have,
$$
\|\boldsymbol{\Theta}_{t+1} - \boldsymbol{\Theta}^*\|^2 - \|\boldsymbol{\Theta}_{t} - \boldsymbol{\Theta}^*\|^2 = -2{\alpha}_t\hat{\nabla}_{{\boldsymbol{\Theta}}_t}(\boldsymbol{\Theta}_{t} - \boldsymbol{\Theta}^*) + {\alpha}^2_t\hat\nabla_{\boldsymbol{\Theta}_{t}^2}.
$$
Taking exception on both sides and use Assumption\ref{ass1}, \ref{ass2} and Lemma \ref{lem:1}, we have
$$
\begin{aligned}
\|\boldsymbol{\Theta}_{t+1} - \boldsymbol{\Theta}^*\|^2 - \|\boldsymbol{\Theta}_{t} - \boldsymbol{\Theta}^*\|^2 & = 
-2{\alpha}_t{\nabla}_{{\boldsymbol{\Theta}}_t}(\boldsymbol{\Theta}_{t} - \boldsymbol{\Theta}^*) + {\alpha}^2_t\mathbb{E}[\hat\nabla_{\boldsymbol{\Theta}_{t}^2}]\\
& \leq -2{\alpha}_t\left[{\mathcal{L}}_t\left({\boldsymbol{\Theta}}_t\right)-{\mathcal{L}}_t\left({\boldsymbol{\Theta}}^*\right)\right] + {\alpha}_t^2\sum_{i=1}^{d}(\mathbb{E}[\hat\nabla_{\boldsymbol{\Theta}_{t,i}^2}])\\
&  \leq -2{\alpha}_t\left[{\mathcal{L}}_t\left({\boldsymbol{\Theta}}_t\right)-{\mathcal{L}}_t\left({\boldsymbol{\Theta}}^*\right)\right] + {\alpha}_t^2d({\sigma}^2 + e^2).
\end{aligned}
$$
Using $\alpha \geq \alpha_t$, we have
$$
\|\boldsymbol{\Theta}_{t+1} - \boldsymbol{\Theta}^*\|^2 - \|\boldsymbol{\Theta}_{t} - \boldsymbol{\Theta}^*\|^2 \leq
-2{\alpha}\left[{\mathcal{L}}_t\left({\boldsymbol{\Theta}}_t\right)-{\mathcal{L}}_t\left({\boldsymbol{\Theta}}^*\right)\right] + {\alpha}^2d({\sigma}^2 + e^2)
$$
Sum up for $t=1, \ldots, T$,
$$
\|\boldsymbol{\Theta}_{T+1} - \boldsymbol{\Theta}^*\|^2 - \|\boldsymbol{\Theta}_{1} - \boldsymbol{\Theta}^*\|^2 \leq
-2\alpha R^{SGD}(T)+ {\alpha}^2 Td({\sigma}^2 + e^2).
$$
We can rearrange the above equation and $\left\|\boldsymbol{\Theta}_n-\boldsymbol{\Theta}_m\right\|_2 \leq D$, 
$$
\begin{aligned}
    R^{SGD}(T) &\leq \frac{\|\boldsymbol{\Theta}_{1} - \boldsymbol{\Theta}^*\|^2 - \|\boldsymbol{\Theta}_{T+1} - \boldsymbol{\Theta}^*\|^2}{2\alpha} + \frac{\alpha Td({\sigma}^2 + e^2)}{2}\\
    & \leq \frac{D^2}{2\alpha} + \frac{\alpha Td({\sigma}^2 + e^2)}{2}
\end{aligned}
$$
\subsection{Theorem 4.5: Convergence of Adam}
\textit{Proof.} The iteration of Adam is
$$
\begin{aligned}
    &\left\{
    \begin{array}{l}
        m_t=\beta_{1,t} \cdot m_{t-1}+\left(1-\beta_{1,t}\right) \cdot \hat{\nabla}_{{\boldsymbol{\Theta}}_t}, \\
        v_t=\beta_2 \cdot v_{t-1}+\left(1-\beta_2\right) \cdot\left(\hat{\nabla}_{{\boldsymbol{\Theta}}_t}\right)^2, \\
        \hat{m}_t=\frac{m_t}{1-\beta_1^t}, \hat{v}_t=\frac{v_t}{1-\beta_2^t} \\
        \boldsymbol{\Theta}_{t+1}=\boldsymbol{\Theta}_t-\frac{\alpha}{\sqrt{\hat{v}}+\epsilon} \cdot \hat{m}_t.
    \end{array}
    \right.
\end{aligned}
$$
Using Lemma \ref{lem:1}, we have,
$$
\mathcal{L}_t\left(\boldsymbol{\Theta}_t\right)-\mathcal{L}_t\left(\boldsymbol{\Theta}^*\right) \leq {\nabla}_{{\boldsymbol{\Theta}}_t}^T\left(\theta_t-\theta^*\right)=\sum_{i=1}^d {\nabla}_{{\boldsymbol{\Theta}}_{t,i}}\left(\boldsymbol{\Theta}_{t, i}-\boldsymbol{\Theta}_{, i}^*\right).
$$
From the above update rules presented, we have
$$
\begin{aligned}
\boldsymbol{\Theta}_{t+1} & =\boldsymbol{\Theta}_t-\alpha_t \widehat{m}_t / \sqrt{\widehat{v}_t} \\
&=\boldsymbol{\Theta}_t-\frac{\alpha_t}{1-\beta_1^t}\left(\frac{\beta_{1, t}}{\sqrt{\widehat{v}_t}} m_{t-1}+\frac{\left(1-\beta_{1, t}\right)}{\sqrt{\widehat{v}_t}} \hat{\nabla}_{{\boldsymbol{\Theta}}_t}\right).    
\end{aligned}
$$
For the $i^{th}$ dimension of the parameter, we subtract the scalar $\boldsymbol{\Theta}_{,i}^*$ and square both sides of the above update rule, we have,
$$
\begin{aligned}
\left(\boldsymbol{\Theta}_{t+1, i}-\boldsymbol{\Theta}_{, i}^*\right)^2=&\left(\boldsymbol{\Theta}_{t, i}-\boldsymbol{\Theta}_{, i}^*\right)^2-\frac{2 \alpha_t}{1-\beta_1^t}\left(\frac{\beta_{1, t}}{\sqrt{\widehat{v}_{t, i}}} m_{t-1, i}+\frac{\left(1-\beta_{1, t}\right)}{\sqrt{\widehat{v}_{t, i}}} \hat{\nabla}_{{\boldsymbol{\Theta}}_{t,i}}\right)\left(\boldsymbol{\Theta}_{t, i}-\boldsymbol{\Theta}_{, i}^*\right)\\
&+\alpha_t^2\left(\frac{\widehat{m}_{t, i}}{\sqrt{\widehat{v}_{t, i}}}\right)^2.
\end{aligned}
$$
We can rearrange the above equation and use Young’s inequality, $ab \leq a^2/2 + b^2/2$. Also, it can be shown that
\begin{equation}
\label{eq29}
\sqrt{\widehat{v}_{t, i}}=\sqrt{\sum_{j=1}^t\left(1-\beta_2\right) \beta_2^{t-j} \hat{\nabla}_{{\boldsymbol{\Theta}}_{j, i}^2}} / \sqrt{1-\beta_2^t} \leq\left\|\hat{\nabla}_{{\boldsymbol{\Theta}}_{1: t, i}}\right\|_2,
\end{equation}
and ${\beta}_{1,t} \leq \beta_1$. Then
$$
\begin{aligned}
\hat{\nabla}_{{\boldsymbol{\Theta}}_{t, i}}\left(\boldsymbol{\Theta}_{t, i}-\boldsymbol{\Theta}_{, i}^*\right)= & \frac{\left(1-\beta_1^t\right) \sqrt{\widehat{v}_{t, i}}}{2 \alpha_t\left(1-\beta_{1, t}\right)}\left(\left(\boldsymbol{\Theta}_{t, i}-\boldsymbol{\Theta}_{, i}^*\right)^2-\left(\boldsymbol{\Theta}_{t+1, i}-\boldsymbol{\Theta}_{, i}^*\right)^2\right) \\
& +\frac{\beta_{1, t}}{\left(1-\beta_{1, t}\right)} \frac{\widehat{v}_{t-1, i}^{\frac{1}{4}}}{\sqrt{\alpha_{t-1}}}\left(\boldsymbol{\Theta}_{, i}^*-\boldsymbol{\Theta}_{t, i}\right) \sqrt{\alpha_{t-1}} \frac{m_{t-1, i}}{\widehat{v}_{t-1, i}^{\frac{1}{4}}}\\
& +\frac{\alpha_t\left(1-\beta_1^t\right) \sqrt{\widehat{v}_{t, i}}}{2\left(1-\beta_{1, t}\right)}\left(\frac{\widehat{m}_{t, i}}{\sqrt{\widehat{v}_{t, i}}}\right)^2 \\
\leq & \frac{1}{2 \alpha_t\left(1-\beta_1\right)}\left(\left(\boldsymbol{\Theta}_{t, i}-\boldsymbol{\Theta}_{, i}^*\right)^2-\left(\boldsymbol{\Theta}_{t+1, i}-\boldsymbol{\Theta}_{, i}^*\right)^2\right) \sqrt{\widehat{v}_{t, i}}\\
&+\frac{\beta_{1, t}}{2 \alpha_{t-1}\left(1-\beta_{1, t}\right)}\left(\boldsymbol{\Theta}_{, i}^*-\boldsymbol{\Theta}_{t, i}\right)^2 \sqrt{\widehat{v}_{t-1, i}} \\
& +\frac{\beta_1 \alpha_{t-1}}{2\left(1-\beta_1\right)} \frac{m_{t-1, i}^2}{\sqrt{\widehat{v}_{t-1, i}}}+\frac{\alpha_t}{2\left(1-\beta_1\right)} \frac{\widehat{m}_{t, i}^2}{\sqrt{\widehat{v}_{t, i}}}.
\end{aligned}
$$
We apply Lemma \ref{lem:3} to the above inequality and sum across all the dimensions for $i \in 1,\dots,d$ and the iterations for $t\in 1,\dots,T$:
$$
\begin{aligned}
\sum_{i=1}^d\sum_{t=1}^T \hat{\nabla}_{{\boldsymbol{\Theta}}_{t, i}}\left(\boldsymbol{\Theta}_{t, i}-\boldsymbol{\Theta}_{, i}^*\right)\leq & \sum_{i=1}^d \frac{1}{2 \alpha\left(1-\beta_1\right)}\left(\boldsymbol{\Theta}_{1, i}-\boldsymbol{\Theta}_{, i}^*\right)^2 \sqrt{\widehat{v}_{1, i}}\\
&+\sum_{i=1}^d \sum_{t=2}^T \frac{1}{2\left(1-\beta_1\right)}\left(\boldsymbol{\boldsymbol{\Theta}}_{t, i}-\boldsymbol{\Theta}_{, i}^*\right)^2\left(\frac{\sqrt{\widehat{v}_{t, i}}}{\alpha_t}-\frac{\sqrt{\widehat{v}_{t-1, i}}}{\alpha_{t-1}}\right) \\
&+\frac{\alpha (1+\beta_1)G_{\infty}}{\left(1-\beta_1\right) \sqrt{1-\beta_2}(1-\gamma)^2} \sum_{i=1}^d\left\|\hat{\nabla}_{{\boldsymbol{\Theta}}_{1: T, i}}\right\|_2 \\
& +\sum_{i=1}^d \sum_{t=1}^T \frac{\beta_{1, t}}{2 \alpha_t\left(1-\beta_{1, t}\right)}\left(\boldsymbol{\Theta}_{, i}^*-\boldsymbol{\Theta}_{t, i}\right)^2 \sqrt{\widehat{v}_{t, i}}
\end{aligned}
$$
From the assumption, $\left\|\boldsymbol{\Theta}_t-\boldsymbol{\Theta}^*\right\|_2 \leq D,\left\|\boldsymbol{\Theta}_m-\boldsymbol{\Theta}_n\right\|_{\infty} \leq D_{\infty}$, we have
$$
\begin{aligned}
\sum_{i=1}^d\sum_{t=1}^T \hat{\nabla}_{{\boldsymbol{\Theta}}_{t, i}}\left(\boldsymbol{\Theta}_{t, i}-\boldsymbol{\Theta}_{, i}^*\right) \leq & \frac{D^2}{2 \alpha\left(1-\beta_1\right)} \sum_{i=1}^d \sqrt{T \widehat{v}_{T, i}}+\frac{D_{\infty}^2}{2 \alpha} \sum_{i=1}^d \sum_{t=1}^T \frac{\beta_{1, t}}{\left(1-\beta_{1, t}\right)} \sqrt{t \widehat{v}_{t, i}}\\
&+\frac{\alpha\left(1+\beta_1\right) G_{\infty}}{\left(1-\beta_1\right) \sqrt{1-\beta_2}(1-\gamma)^2}. \sum_{i=1}^d\left\|\hat{\nabla}_{{\boldsymbol{\Theta}}_{1: T, i}}\right\|_2
\end{aligned}
$$
We apply Eq. \ref{eq29} to the above inequality, we have
$$
\begin{aligned}
\sum_{i=1}^d\sum_{t=1}^T \hat{\nabla}_{{\boldsymbol{\Theta}}_{t, i}}\left(\boldsymbol{\Theta}_{t, i}-\boldsymbol{\Theta}_{, i}^*\right) \leq &\frac{D^2\sqrt{T}}{2 \alpha\left(1-\beta_1\right)} \sum_{i=1}^d \|\hat{\nabla}_{{\boldsymbol{\Theta}}_{1: T, i}}\|_2+\frac{D_{\infty}^2}{2 \alpha} \sum_{i=1}^d \sum_{t=1}^T \frac{\beta_{1, t}\sqrt{t}}{\left(1-\beta_{1, t}\right)}  \|\hat{\nabla}_{{\boldsymbol{\Theta}}_{1: t, i}}\|_2\\
&+\frac{\alpha\left(1+\beta_1\right) G_{\infty}}{\left(1-\beta_1\right) \sqrt{1-\beta_2}(1-\gamma)^2} \sum_{i=1}^d\left\|\hat{\nabla}_{{\boldsymbol{\Theta}}_{1: T, i}}\right\|_2
\end{aligned}
$$
Take expectation on both sides of the above inequality and apply Lemma \ref{lem:4}, Assumption \ref{ass1},
$$
\begin{aligned}
\sum_{i=1}^d\sum_{t=1}^T {\nabla}_{{\boldsymbol{\Theta}}_{t, i}}\left(\boldsymbol{\Theta}_{t, i}-\boldsymbol{\Theta}_{, i}^*\right) \leq &\frac{D^2\sqrt{T}}{2 \alpha\left(1-\beta_1\right)} \sum_{i=1}^d \mathbb{E}[\|\hat{\nabla}_{{\boldsymbol{\Theta}}_{1: T, i}}\|_2]\\
&+\frac{\alpha\left(1+\beta_1\right) G_{\infty}}{\left(1-\beta_1\right) \sqrt{1-\beta_2}(1-\gamma)^2} \sum_{i=1}^d\mathbb{E}[|\hat{\nabla}_{{\boldsymbol{\Theta}}_{1: T, i}}\|_2]\\
&+\frac{D_{\infty}^2}{2 \alpha} \sum_{i=1}^d \sum_{t=1}^T \frac{\beta_{1, t}\sqrt{t}}{\left(1-\beta_{1, t}\right)}  \mathbb{E}[\|\hat{\nabla}_{{\boldsymbol{\Theta}}_{1: t, i}}\|_2]\\
\leq & \frac{D^2\sqrt{T}}{2 \alpha\left(1-\beta_1\right)} \sum_{i=1}^d \sqrt{\mathbb{E}[\|\hat{\nabla}_{{\boldsymbol{\Theta}}_{1: T, i}}\|_2^2]}\\
&+\frac{\alpha\left(1+\beta_1\right) G_{\infty}}{\left(1-\beta_1\right) \sqrt{1-\beta_2}(1-\gamma)^2} \sum_{i=1}^d\sqrt{\mathbb{E}[|\hat{\nabla}_{{\boldsymbol{\Theta}}_{1: T, i}}\|_2^2]}\\
&+\frac{D_{\infty}^2}{2 \alpha} \sum_{i=1}^d \sum_{t=1}^T \frac{\beta_{1, t}\sqrt{t}}{\left(1-\beta_{1, t}\right)}  \sqrt{\mathbb{E}[\|\hat{\nabla}_{{\boldsymbol{\Theta}}_{1: t, i}}\|_2^2]}\\
\leq&\frac{D^2{T}}{2 \alpha\left(1-\beta_1\right)} \sum_{i=1}^d \sqrt{\sigma^2+e^2}+\frac{\alpha\left(1+\beta_1\right) G_{\infty}\sqrt{T}}{\left(1-\beta_1\right) \sqrt{1-\beta_2}(1-\gamma)^2} \sum_{i=1}^d\sqrt{\sigma^2+e^2}\\
&+\frac{D_{\infty}^2}{2 \alpha} \sum_{i=1}^d \sum_{t=1}^T \frac{\beta_{1, t}{t}}{\left(1-\beta_{1, t}\right)}  \sqrt{\sigma^2+e^2}.
\end{aligned}
$$
We can use arithmetic geometric series upper bound for the last term:
$$
\begin{aligned}
\sum_{t=1}^T \frac{\beta_{1, t}}{\left(1-\beta_{1, t}\right)} {t} & \leq \sum_{t=1}^T \frac{1}{\left(1-\beta_1\right)} \lambda^{t-1} {t} 
 \leq \frac{1}{\left(1-\beta_1\right)(1-\lambda)^2}
\end{aligned}
$$
Therefore, we have the following regret bound:
$$
\begin{aligned}
R(T)\leq &\sum_{i=1}^d\sum_{t=1}^T {\nabla}_{{\boldsymbol{\Theta}}_{t, i}}\left(\boldsymbol{\Theta}_{t, i}-\boldsymbol{\Theta}_{, i}^*\right)\\
\leq &\frac{((1-\lambda)^2D^2{T} + D_{\infty}^2)d}{2 \alpha\left(1-\beta_1\right)(1-\lambda)^2}  \sqrt{\sigma^2+e^2}+\frac{\alpha\left(1+\beta_1\right) G_{\infty}\sqrt{T}d}{\left(1-\beta_1\right) \sqrt{1-\beta_2}(1-\gamma)^2} \sqrt{\sigma^2+e^2}\\
\end{aligned}
$$
\section{Variance of Specific Quantizers}\label{appendix:variance}
\begin{proposition}
(Variance of stochastic rounding) For any number ${X} \in \mathbb{R}$, $\operatorname{Var}[\operatorname{SR}({X})] \leq \frac{1}{4}$.
\end{proposition}
\textit{Proof.} For any real number $X$, let $p:=X-\lfloor X\rfloor \in[0,1)$, then
$$
\begin{aligned}
& \operatorname{Var}[\operatorname{SR}(X)]=\mathbb{E}[\operatorname{SR}(X)-X]^2=p(\lceil X\rceil-X)^2+(1-p)(\lfloor X\rfloor-X)^2 \\
= & p(1-p)^2+p^2(1-p)=p(1-p)(1-p+p)=p(1-p) \leq \frac{1}{4} .
\end{aligned}
$$
\subsection{Per-sample Quantizer}
Given an activation gradient $\hat{\nabla}_{\mathbf{H}^{(l)}}$, its per-sample quantization is:
$$
Q_g(\hat{\nabla}_{\mathbf{H}^{(l)}_{i,j}})=\operatorname{SR}(B(\hat{\nabla}_{\mathbf{H}^{(l)}_{i,j}}-Z_i)/R_i)R_i/B+Z_i,
$$
where apply different ranges $R_i$ and zero points $Z_i$ for each sample of the gradient.
When $\mathbf{S}=\operatorname{diag}\left\{\frac{R_1}{B}, \ldots, \frac{R_N}{B}\right\}$, we have
$$
\begin{aligned}
\operatorname{Var}\left[Q_g\left(\hat{\nabla}_{\mathbf{H}^{(l)}}\right)\right]& =\operatorname{Var}\left[\mathbf{S}\operatorname{SR}\left((\mathbf{S}^{{-1}}\left(\hat{\nabla}_{\mathbf{H}^{(l)}}-\mathbf{1}\mathbf{Z}\right)\right) / +\mathbf{1}\mathbf{Z}\right] \\
& \leq \sum_{i=1}^N\sum_{j=1}^{D^{(l)}}\operatorname{Var}[\frac{R_i}{B}\operatorname{SR}(\frac{B}{R_i}(\hat{\nabla}_{\mathbf{H}^{(l)}_{i,j}}-Z_i))+Z_i)]\\
&=\sum_{i=1}^N\sum_{j=1}^{D^{(l)}}\frac{R_i^2}{B^2}\operatorname{Var}[\operatorname{SR}(\frac{B}{R_i}(\hat{\nabla}_{\mathbf{H}^{(l)}_{i,j}}-Z_i))]\\
&\leq\frac{D^{(l)}}{4B^2}\sum_{i=1}^N R_i^2.
\end{aligned}
$$
\subsection{Per-sample Quantizer with AGP}

Place the groups with the largest range in the first $N/b$ rows, and let the range of these groups be denoted by {$R_1, \dots, R_{N/b}$}, groups in the remaining rows are denoted by {$r_{N/b+1}, \dots, r_N$}. We assume that $r/R\approx0$. 
$$
Q_g(\hat{\nabla}_{\mathbf{H}^{(l)}})=({\mathbf{M}}\mathbf{S})\operatorname{SR}\left({(\mathbf{M}\mathbf{S}^{}})^{-1}\left(\mathbf{M}\hat{\nabla}_{\mathbf{H}^{(l)}}-\mathbf{M}\mathbf{Z}^{}\right)\right) +\mathbf{M}\mathbf{Z}^{},
$$
where ${\mathbf{M}}=\operatorname{diag}\left(\frac{m_1}{p_1}, \ldots, \frac{m_{N}}{p_{N}}\right)$, $p_i=\frac{NR_i}{bR_{total}}$, $R_{total}=\sum_{i=1}^{N}R_i$ and $m_i \sim \operatorname{Bern}\left(p_i\right)$. To simplify the problem, we assume that $R_1 \approx R_2 \dots\approx R_{N/b}$. And we use 
$r/R\approx0$, then $p\approx\{1,\dots,0\}$. In other words, for the first $\frac{N}{b}$ rows, $m=1$, and 0 otherwise. We substitute it into the above equation and prune the groups with smaller ranges, 
$$
Q_g(\hat{\nabla}_{\mathbf{H}^{(l)}})=\mathbf{S}^{}_{1:\frac{N}{b}, 1:\frac{N}{b}}\operatorname{SR}\left({(\mathbf{S}^{}_{1:\frac{N}{b}, 1:\frac{N}{b}}})^{-1}\left(\hat{\nabla}_{\mathbf{H}^{(l)}_{1:\frac{N}{b}}}-\mathbf{1}\mathbf{Z}^{}_{1:\frac{N}{b}}\right)\right) +\mathbf{1}\mathbf{Z}^{}_{1:\frac{N}{b}}.
$$
Then we have,
$$
\begin{aligned}
\operatorname{Var}\left[Q_g\left(\hat{\nabla}_{\mathbf{H}^{(l)}}\right)\right]& \leq \sum_{i=1}^{N/b}\sum_{j=1}^{D^{(l)}}\operatorname{Var}[\frac{R_i}{B}\operatorname{SR}(\frac{B}{R_i}(\hat{\nabla}_{\mathbf{H}^{(l)}_{i,j}}-Z_i))+Z_i)]\\
&=\sum_{i=1}^{N/b}\sum_{j=1}^{D^{(l)}}\frac{R_i^2}{B^2}\operatorname{Var}[\operatorname{SR}(\frac{B}{R_i}(\hat{\nabla}_{\mathbf{H}^{(l)}_{i,j}}-{Z}_i))]\\
&\leq\frac{D^{(l)}}{4B^2}\sum_{i=1}^{N/b} R_i^2.
\end{aligned}
$$
For 1-bit quantizers, the variance of PSQ is $\frac{D^{(l)}}{4}(\sum_{i=1}^{N/b}R_i^2+\sum_{i=N/b+1}^{N}r_i^2)$. It is clear that
$$
\frac{D^{(l)}}{4B^2}\sum_{i=1}^{N/b} R_i^2\leq\frac{D^{(l)}}{4}(\sum_{i=1}^{N/b}R_i^2+\sum_{i=N/b+1}^{N}r_i^2).
$$
\section{Implementation details}\label{appendix:implementation}
For simplicity, we use $\overline{\nabla}_{\mathbf{H}^{(l)}_{PSQ}}$ and $\overline{\nabla}_{\mathbf{H}^{(l)}_{PCQ}}$ to denote the low-bit versions quantized from ${\mathbf{M}}_{PSQ}\hat{\nabla}_{\mathbf{H}^{(l)}}$ and $\hat{\nabla}_{\mathbf{H}^{(l)}}{\mathbf{M}}_{PCQ}$. Extend this decomposition operation to the entire gradient tensor:
$$
\begin{aligned}
    &\left\{
    \begin{array}{l}
        \hat{\nabla}_{\mathbf{H}^{(l-1)}}=\sum_{i=1}^{b}2^{(i-1)}(\mathbf{S}_{PSQ}^{}(\overline{\nabla}_{\mathbf{H}^{(l)}_{PSQ}})^i \operatorname{sign}({\boldsymbol{\Theta}}^{(l)^{\top}})), \\
        \hat{\nabla}_{\boldsymbol{\Theta}^{(l)}}=\sum_{i=1}^{b}2^{(i-1)}(\operatorname{sign}({\mathbf{H}}^{(l-1)^{\top}}) (\overline{\nabla}_{\mathbf{H}^{(l)}_{PCQ}})^{i}\mathbf{S}_{PCQ}^{}),
    \end{array}
    \right.
\end{aligned}
$$
where both $(\overline{\nabla}_{\mathbf{H}^{(l)}_{PSQ}})^i$ and $(\overline{\nabla}_{\mathbf{H}^{(l)}_{PCQ}})^i$ represent the 1-bit tensors obtained after the decomposition. This transformation enables the implementation of most operations in backward propagation using XNOR and bit counting.\par
We implemented our method as a lightweight library in PyTorch. For binary matrix multiplication and some auxiliary operations, we implemented them using C++. In Alg. \ref{alg1}, we illustrate the process of forward and backward propagation for quantized fully connected layers. For simplicity, certain details, such as bias terms, quantization zero points, and the decomposition operations on gradient tensors, are omitted here. The entire process primarily consists of five components: quantization (9), encoding(4-5, 10), low-bit multiplication (6, 11), pruning (8), and dequantization (12).\par

\begin{algorithm}
   \caption{Linear Layer Forward and Backward Propagation}
   \label{alg1}
   \begin{minipage}{\columnwidth}
      \begin{algorithmic}[1]
         \State {\bfseries Input:} Input $\mathbf{H}^{(l-1)}$, Weight $\boldsymbol{\Theta}^{(l)}$, Gradient of Loss ${\nabla}_{\mathbf{H}^{(l)}}$
         \State {\bfseries Output:} Output $\mathbf{H}^{(l)}$, Gradient of Weight ${\nabla}_{\boldsymbol{\Theta}^{(l)}}$, Gradient of Input ${\nabla}_{\mathbf{H}^{(l-1)}}$

     \State // Forward Propagation
     \State Encode Weight: $\tilde{\mathbf{H}}^{(l-1)}=\operatorname{row\_encoder}({\mathbf{H}}^{(l-1)})$
     \State Encode Input: $\tilde{\boldsymbol{\Theta}}^{(l)}=\operatorname{column\_encoder}({\boldsymbol{\Theta}}^{(l)})$
         \State Compute Output: $\mathbf{H}^{(l)} = \tilde{\mathbf{H}}^{(l-1)} \tilde{\boldsymbol{\Theta}}^{(l)}$

         \State // Backward Propagation
         \State Pruning: ${\nabla}_{\mathbf{H}^{(l)}_{PSQ}},{\nabla}_{\mathbf{H}^{(l)}_{PCQ}}=\operatorname{pruner}({\nabla}_{\mathbf{H}^{(l)}})$
         \State Quantization: $\overline{\nabla}_{\mathbf{H}_{PSQ}^{(l)}},\mathbf{S}_{PSQ}^{(l)}=\operatorname{PSQ}({\nabla}_{\mathbf{H}^{(l-1)}_{PSQ}}),\newline
         \overline{\nabla}_{\mathbf{H}_{PCQ}^{(l)}},\mathbf{S}_{PCQ}^{(l)}=\operatorname{PCQ}({\nabla}_{\mathbf{H}^{(l-1)}_{PCQ}})$
         \State Encode Gradient: $\overline{\nabla}_{\mathbf{H}_{PSQ}^{(l)}}=\operatorname{row\_encoder}(\overline{\nabla}_{\mathbf{H}_{PSQ}^{(l)}}),\newline
         \overline{\nabla}_{\mathbf{H}_{PCQ}^{(l)}}=\operatorname{column\_encoder}(\overline{\nabla}_{\mathbf{H}_{PCQ}^{(l)}})$
         \State Compute Gradient: $\overline{\nabla}_{\boldsymbol{\Theta}^{(l)}} = \tilde{\mathbf{H}}^{(l-1)^{\top}} \overline{\nabla}_{\mathbf{H}_{PCQ}^{(l)}},\newline
         \overline{\nabla}_{\mathbf{H}^{(l-1)}}=\overline{\nabla}_{\mathbf{H}_{PSQ}^{(l)}}\tilde{\boldsymbol{\Theta}}^{(l)^{\top}}$

        \State Dequantization: $\hat{\nabla}_{\boldsymbol{\Theta}^{(l)}}=\overline{\nabla}_{\boldsymbol{\Theta}^{(l)}}\mathbf{S}_{PCQ}^{(l)}$, $\hat{\nabla}_{\mathbf{H}^{(l-1)}}=\mathbf{S}_{PSQ}^{(l)}\overline{\nabla}_{\mathbf{H}^{(l-1)}}$
         \State // Update Parameters
         \State Update Weight: $\mathbf{W} \leftarrow \mathbf{W} - \alpha \hat{\nabla}_{\boldsymbol{\Theta}^{(l)}}$
      \end{algorithmic}
   \end{minipage}
\end{algorithm}

\textbf{Encoder} is a functional component that encodes multiple integers with values of 1 or -1 into a smaller set of elements, facilitating subsequent XNOR operations. Taking Row\_Encoder as an example, its primary form is illustrated in  Alg. \ref{alg2}, the case where the number of columns is not divisible by $b$ has been overlooked.\par
\begin{algorithm}
\caption{Row\_Encoder}
\label{alg2}
\begin{algorithmic}[1]
 \State {\bfseries Input:} Input $\mathbf{H} \in \mathbb{R}^{N \times D}$, Bits $b$
 \State {\bfseries Output:} Output $\mathbf{H}_e \in \mathbb{R}^{N \times \lfloor(1 + (D - 1)/b)\rfloor}$
\For{$i \gets 1$ to $N$}
    \For{$j \gets 1$ to $\lfloor(1 + (D - 1)/b)\rfloor$}
        \State $\mathbf{H}_{i,j}=0$
        \For{$k \gets 1$ to $b$}
            \State $s = (\mathbf{H}>0)$
            \State $\mathbf{H}_{i,j}=(\mathbf{H}_{i,j}<<1)\|s$
        \EndFor
    \EndFor
\EndFor
\end{algorithmic}
\end{algorithm}
\textbf{Binary multiplication} is the crucial operation. In our approach, both forward and backward propagation are implemented through binary multiplication. For example, 
For two vectors, $\mathbf{X}_1$
  and $\mathbf{X}_2$, each of length 32, encode them into two unsigned 32-bit integers, $x_1$ and $x_2$. The multiplication of the two is implemented as follows:
  $$
\operatorname{SUM}(\mathbf{X}_1 \odot \mathbf{X}_2)=\operatorname{bitcount}(\operatorname{XNOR}(x1,x2))<<1-32
$$
 where the dot product of two vectors, each of length 32, is efficiently replaced by a single bitcount and XNOR operation, effectively reducing energy consumption and time overhead. 
However, it is worth noting that an unbiased quantizer maps data to 0 or 1, rather than -1 or 1. Therefore, some conversion is required. For $\mathbf{X}_1 \in {\{1,-1\}}^n$, $\mathbf{X}_2 = \operatorname{ReLU}(\mathbf{X}_1)$, it is clear that $(S/2)\mathbf{X}_1+Z+(S/2)=S\mathbf{X}_2+Z$. Therefore, some adjustments are needed: a straightforward modification of the scaling factor $S$ and zero point $Z$ is sufficient to achieve the transformation. This transformation requires only one multiplication and one addition for the scaling factor and zero point, thus incurring minimal overhead.\par
For convolutional layers, direct matrix multiplication is not feasible. To facilitate subsequent operations, an unfolder is performed on the convolutional layer before matrix multiplication. After computation, the standard form is restored through folder operations. For example, We perform a convolution operation between the input $\mathbf{X} \in \mathbb{R}^{N \times C \times H \times W}$and parameters ${\boldsymbol{\Theta}} \in \mathbb{R}^{D \times C \times K \times K}$ to obtain the output $\mathbf{Y} \in \mathbb{R}^{N \times D \times H \times W}$.\par
\begin{figure*}[t]
\begin{center}
\centerline{\includegraphics[width=\textwidth]{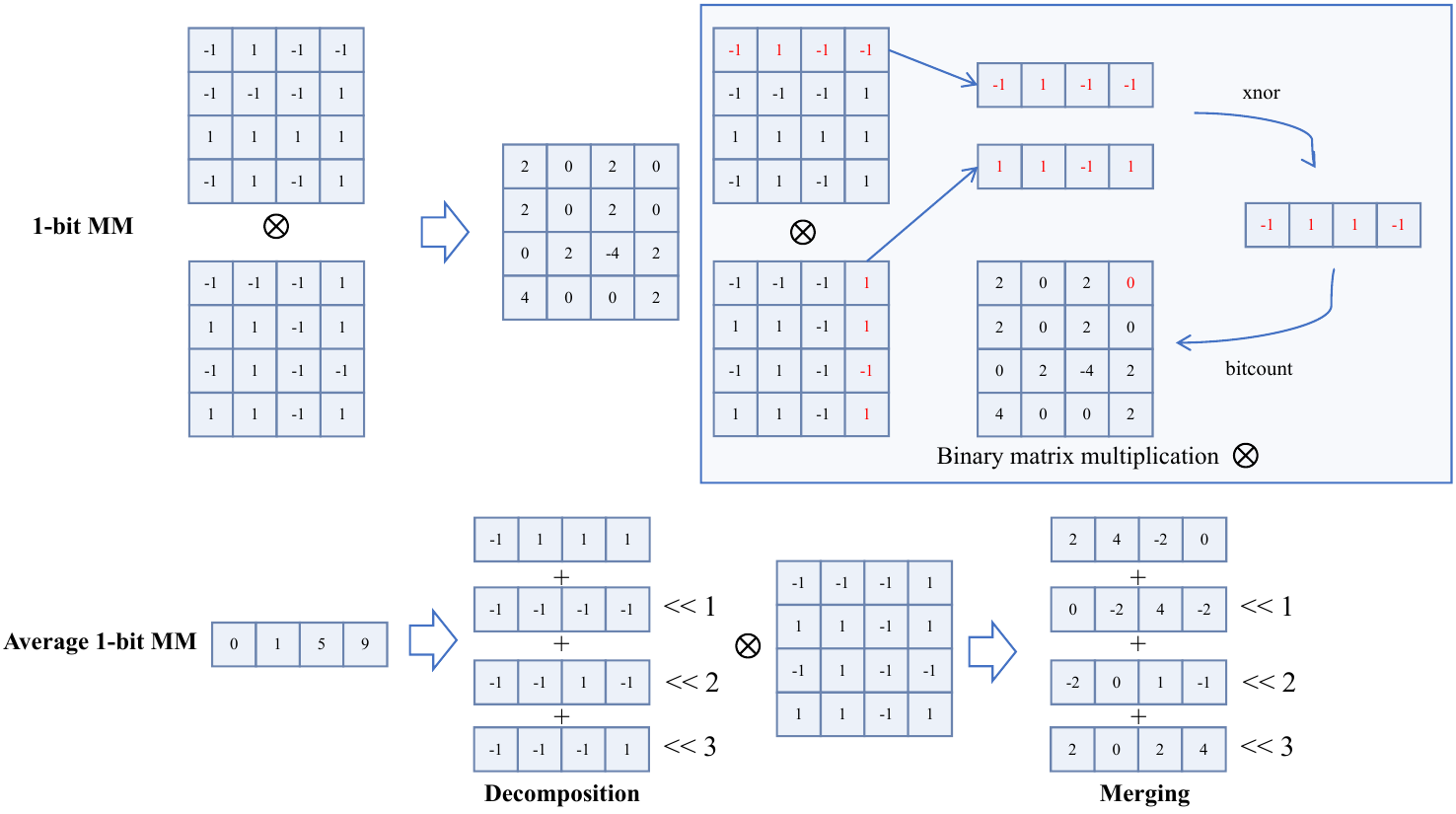}}
\caption{The calculation process of average 1-bit matrix multiplication and 1-bit matrix multiplication.}
\label{1-bitmm}
\end{center}
\vskip -0.2in
\end{figure*}

\textbf{1-bit  Matrix Multiplication vs. Average 1-bit Matrix Multiplication.} Given that our method is not a true 1-bit FQT but rather an average 1-bit FQT, we analyze the computational differences between the two approaches. In the case of b=4, the "average 1-bit matrix multiplication" first prunes a matrix to one-fourth of its original size. The remaining elements in this pruned matrix have a bit-width of 4 bits, ensuring an average bit-width of 1 bit. Subsequently, this pruned matrix is multiplied with another true 1-bit matrix. The computation process is as follows: based on the lossless decomposition described in the paper, the 4-bit matrix is decomposed into four equally sized 1-bit matrices. Subsequently, each of these four matrices undergoes true 1-bit matrix multiplication with the genuine 1-bit matrix. Finally, the four resulting matrices are merged together (Fig. \ref{1-bitmm}). The operations involved in 1-bit matrix multiplication of the four submatrices are equivalent to those used in true 1-bit matrix multiplication, so no further analysis is required. For the decomposition operation, it can be accomplished through bitwise AND and shift operations. As for the final merging operation, it can be achieved using shift operations and integer addition. Moreover, the number of computations involved in both operations is at the level of the matrix size. In summary, while average 1-bit matrix multiplication requires additional decomposition and merging operations compared to true 1-bit matrix multiplication, these operations can be achieved using bitwise AND, shift, and integer addition operations. The overhead associated with these operations is relatively low.\par
\textbf{Unfolder} treats each element involved in element-wise multiplication within the kernel as a row, and the number of times the window slides as columns, the unfolded input and parameters transform into $\mathbf{X}_u \in \mathbb{R}^{NHW \times CK^2}$, ${\boldsymbol{\Theta}}_u \in \mathbb{R}^{D \times CK^2}$. Finally, we need to restore the output $\mathbf{Y}_u \in \mathbb{R}^{NHW \times D}$ to its standard state.\par
\textbf{Folder} is the inverse operation of Unfolder, designed to restore the gradients of both the input and parameters $\nabla_{\mathbf{X}_u} \in \mathbb{R}^{NHW \times CK^2}$,  $\nabla_{\boldsymbol{\Theta}_u} \in \mathbb{R}^{D \times CK^2}$ to their standard states.
\section{Experimental Details}\label{appendix:experiment}
\subsection{Gradient distribution}
\begin{figure*}[t]
  \centering
  \subfigure[0]{
    \includegraphics[width=0.15\columnwidth]{he/epoch0.pdf}
    \label{fig6:1}
  }
  \hspace{0.0\columnwidth}  
  \subfigure[15]{
    \includegraphics[width=0.15\textwidth]{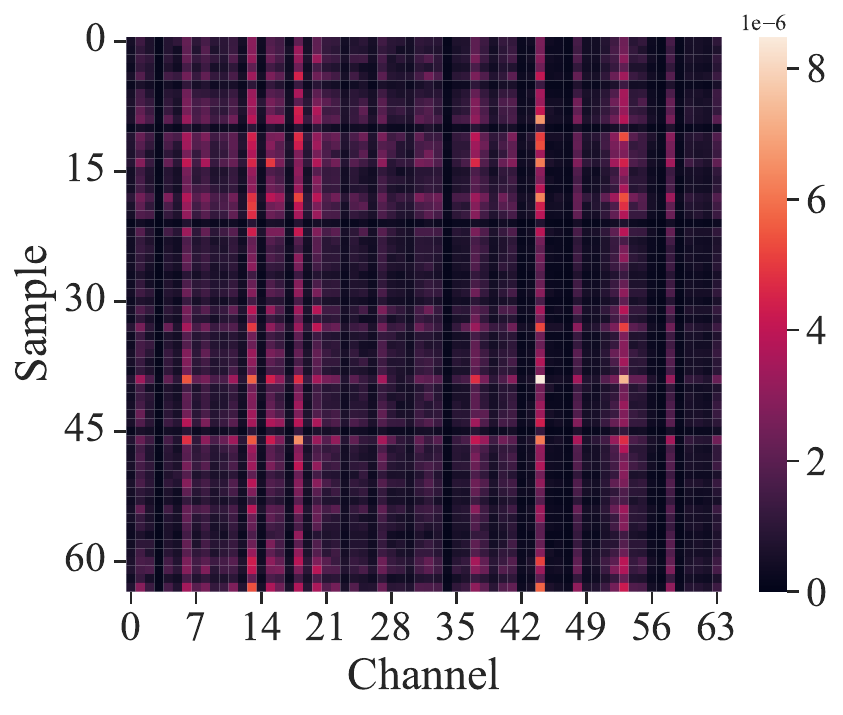}
    \label{fig6:2}
  }
  \hspace{0.0\columnwidth}  
  \subfigure[30]{
    \includegraphics[width=0.15\textwidth]{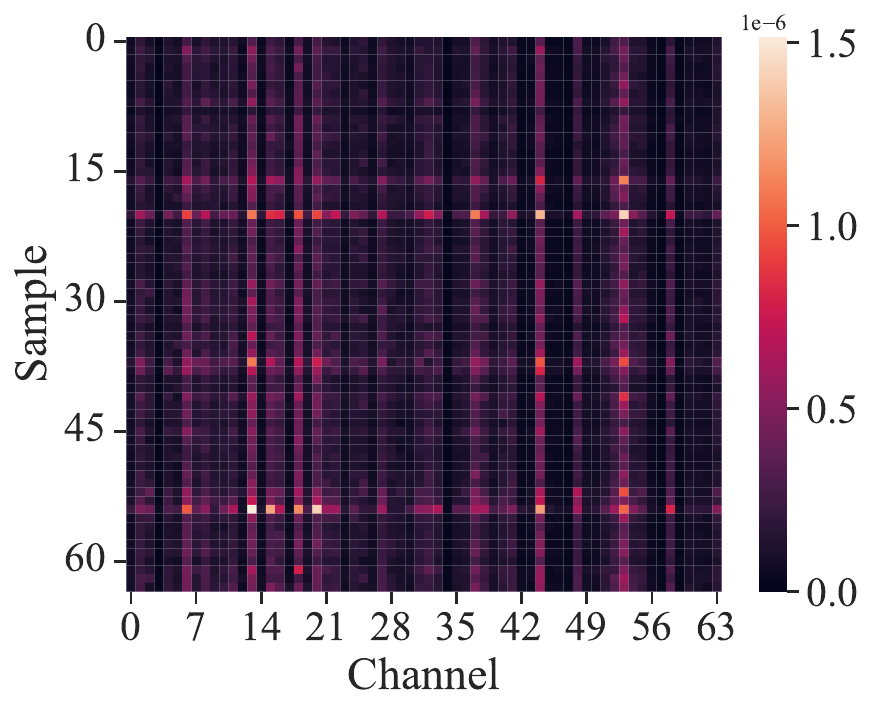}
    \label{fig6:3}
  }
  \hspace{0.0\columnwidth}  
  \subfigure[45]{
    \includegraphics[width=0.15\textwidth]{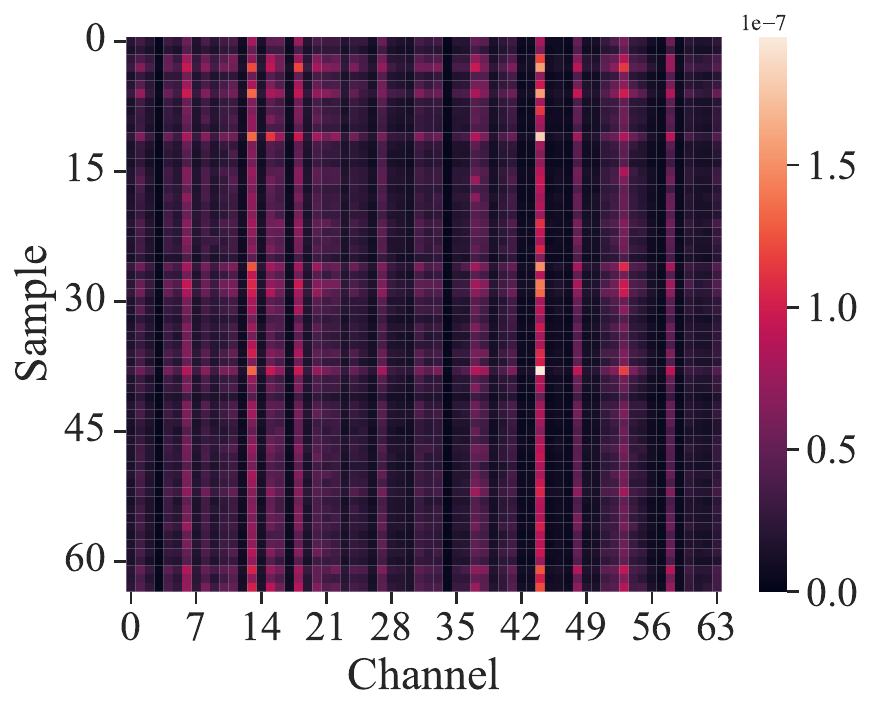}
    \label{fig6:4}
  }
  \hspace{0.0\columnwidth}  
  \subfigure[60]{
    \includegraphics[width=0.15\textwidth]{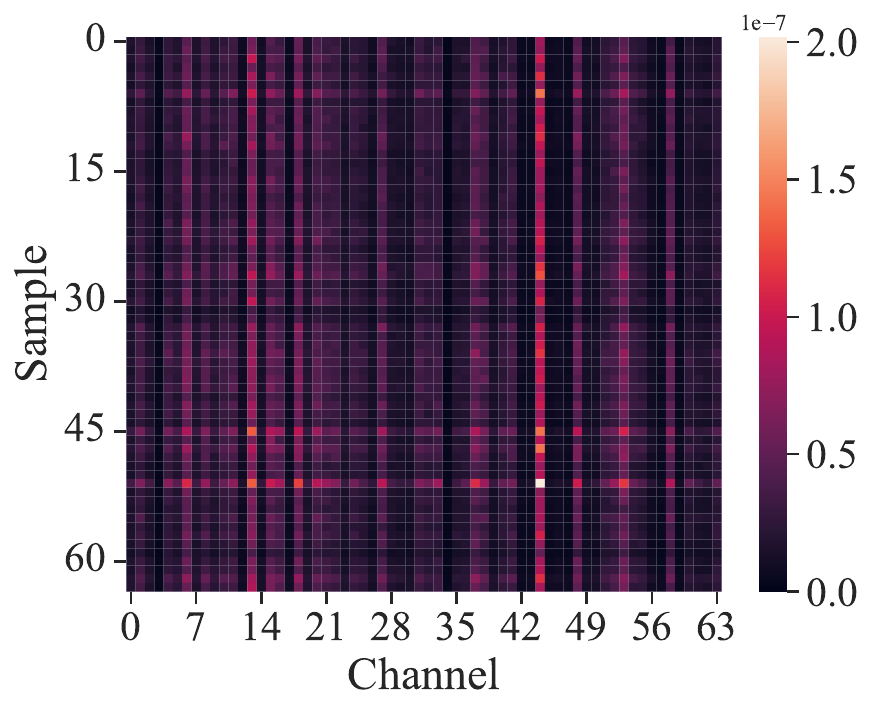}
    \label{fig6:5}
  }
  \hspace{0.0\columnwidth}  
  \subfigure[]{
    \includegraphics[width=0.15\textwidth]{Styles/picdotc-2.pdf}
    \label{fig6:6}
  }
  \hspace{0.0\columnwidth}  
  \subfigure[]{
    \includegraphics[width=0.15\textwidth]{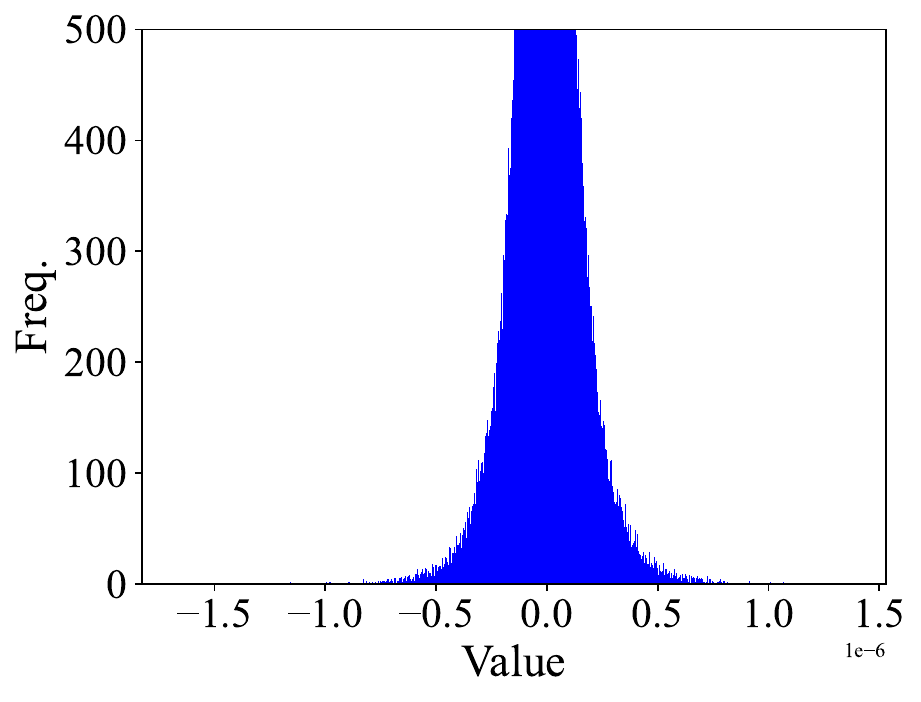}
    \label{fig6:7}
  }
  \hspace{0.0\columnwidth}  
  \subfigure[]{
    \includegraphics[width=0.15\textwidth]{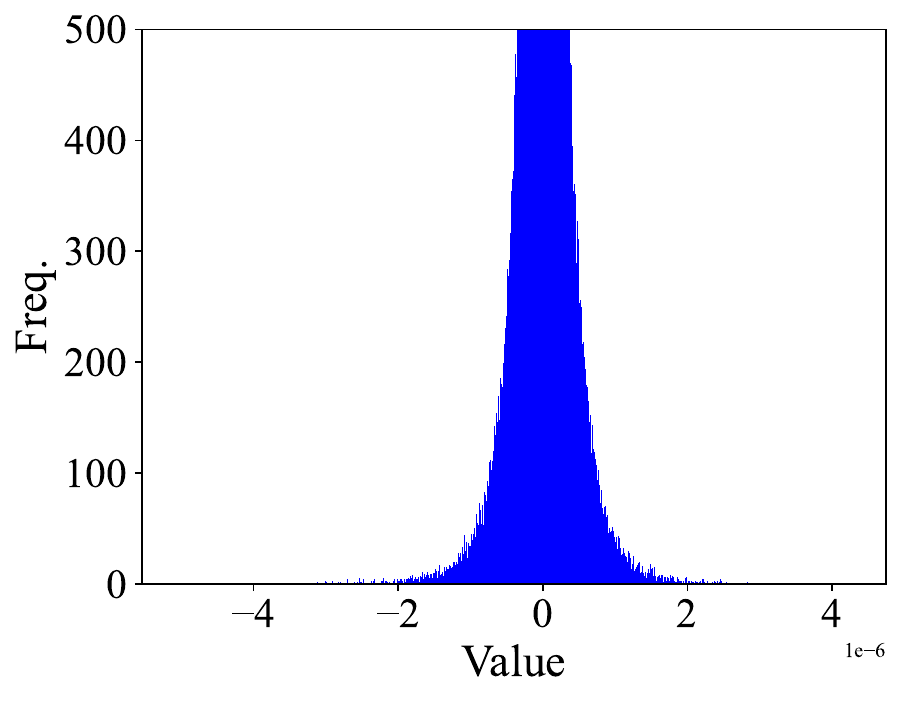}
        \label{fig6:8}
  }
  \hspace{0.0\columnwidth}  
  \subfigure[]{
    \includegraphics[width=0.15\textwidth]{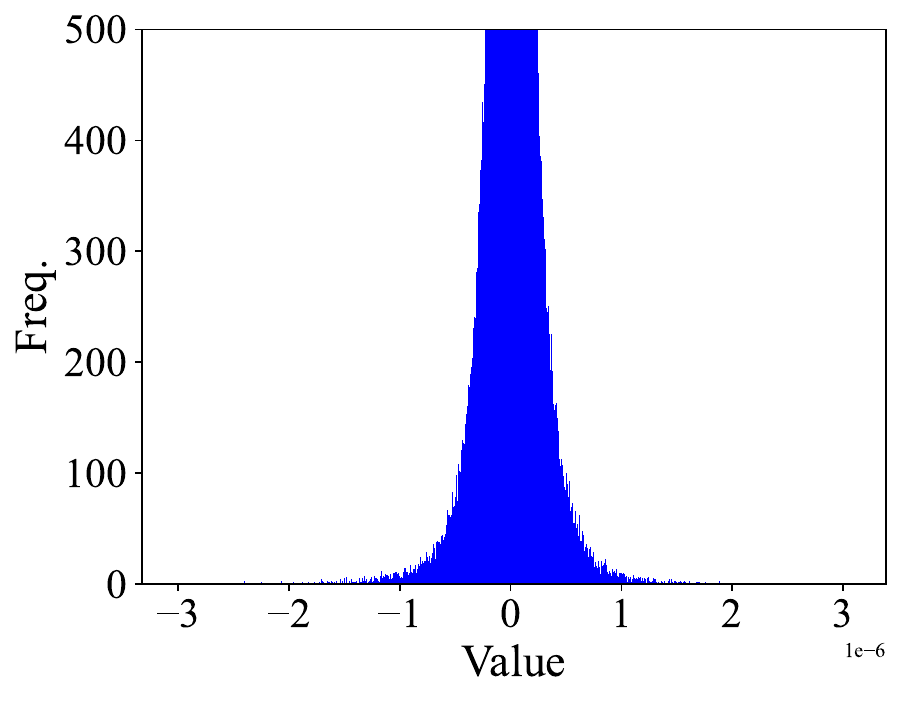}
        \label{fig6:9}
  }
  \hspace{0.0\columnwidth}  
  \subfigure[]{
    \includegraphics[width=0.15\textwidth]{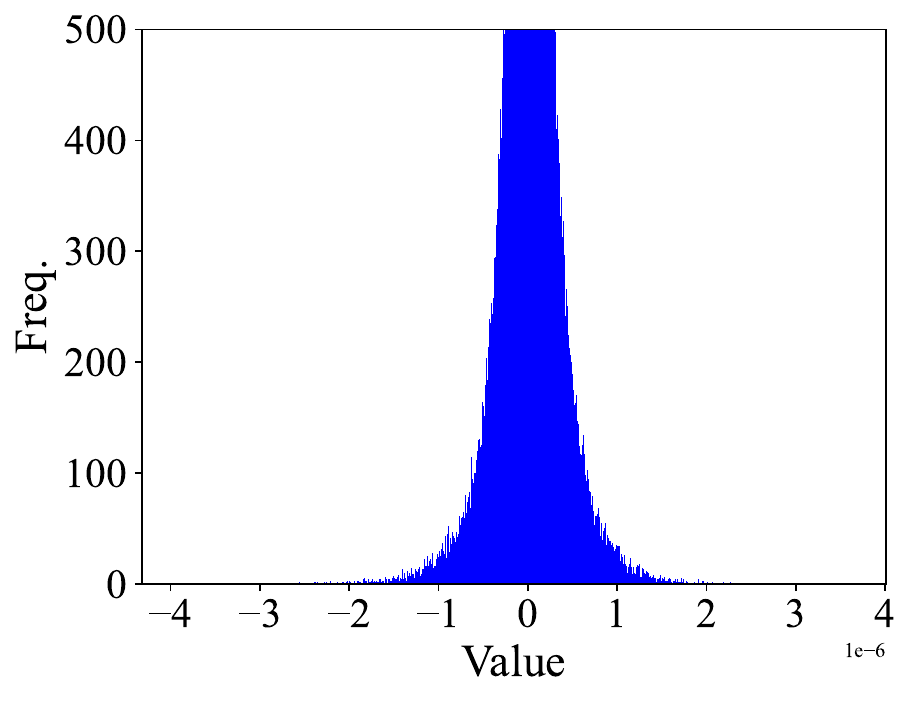}
    \label{fig6:10}
  }
    \caption{Heterogeneity in a ResNet18’s gradients. (a-e) Heatmap of the per-group range at the conv2.1.2 layer under different epochs; (f-j) Histogram of the gradient groups (samples) at the same layer.}
\label{fig6}

\end{figure*}

From Fig. \ref{fig6}, it can be observed that the gradient of the activation exhibits a pattern across different epochs: the ranges of groups (both samples and dimensions) are highly uneven. Some groups have large ranges, while others have small ranges. Although we have presented results for a single batch, the same pattern persists across the remaining batches.
\subsection{Experimental settings}
\textbf{Classification task:} The training process is divided into two stages: initially undergoing quantization-aware training on ImageNet and subsequently undergoing FQT on various downstream datasets. The first stage: the initial learning rate was set to $10^{-3}$ and the weight decay to $10^{-5}$, following \cite{bulat2019xnor}, the optimizer is Adam and use a consine learning rate schedule. We train for 90 epochs. The second stage: for all datasets, the initial learning rate for fully connected layers is set to $10^{-3}$. For portions of the network that have been previously trained, the learning rate is set to $10^{-5}$, except for car dataset \cite{krause20133d} where it is set to $10^{-4}$. The optimizer settings are the same as the first stage. We train for 60 epochs. The batch size was assigned to be 128. We measured training latency on CPUs, but to expedite the acquisition of accuracy statistics, we simulated the training results on 4 NVIDIA RTX A4000 GPUs. Due to limited resources on terminal devices, we utilized a smaller batch size of 64. We followed the configuration of \cite{bulat2019xnor} by excluding quantization for sensitive layers, such as the first and last layers, as well as skip connections in residual networks, in addition to batch normalization (BN) and ReLU layers.\par
\textbf{Detection task:} We evaluate our method on a simple transfer learning task to assess its effectiveness on object detection models, specifically transferring from high-resolution object detection to low-resolution object detection. The training process is divided into two stages: initially undergoing quantization-aware training on the PASCAL VOC 2007 and VOC 2012 trainval sets with a resolution of (600*600) pixels, followed by FQT training on the same dataset with a resolution of (300*300) pixels. The first stage: We followed all the settings of BiDet \cite{wang2020bidet}, including the quantization methods for both weights and activation values and training configurations. The batch size was assigned to be 32, and the Adam optimizer was applied. The learning rate started from $10^{-3}$ and dropped during training every 6 epochs by a factor of 10. We train for 20 epochs. The second stage: the initial learning rate is $10^{-5}$, the training epoch is 5 and the others are the same.\par
\textbf{NLP tasks:} We conduct experiments to validate the effectiveness of our proposed 1-bit FQT on $\text{BERT}_{\text{BASE}}$(12 hidden layers) and the GLUE benchmark \cite{wang2018glue} which consists of nine basic language tasks. We use the standard metrics for each GLUE task to evaluate our method. We use Spearman Correlation for STS-B, Mathews Correlation Coefficient for CoLA and classification accuracy for the rest tasks. As for MNLI task, we report the accuracy on both in-domain evaluation MNLI-match (MNLI-m) and cross-domain evaluation MNLI-mismatch (MNLI-mm). We exclude WNLI task as \cite{qin2022bibert}. We utilized BiBERT\cite{qin2022bibert} as our binarized model, which is derived by directly binarizing a full-precision one. Subsequently, we fine-tune this binarized model using both full-precision gradients (QAT) and 1-bit gradients (Ours). We follow \cite{qin2022bibert} by excluding quantization for classifier, position embedding layer, and token type embedding layer. We use the Adam as our optimizer. The training settings are also the same as \cite{qin2022bibert}.


		
		

\begin{table*}[t]
\vskip -0.1in
\caption{Experimental results on multiple downstream datasets. ``(W, A, G)" denote the bitwidth of weight, activations, and gradients, respectively.}

\small
	\centering

    \setlength{\tabcolsep}{1mm}
	\begin{tabular}{l c c c c c c c r}
		\toprule
		
		\multirow{2}{*}{Method} & {Precision} & \multicolumn{7}{c}{Accuracy(\%)} \\
        \cline{3-9}
        &{(W, A, G)}& CIFAR-10 & CIFAR-100 & Flowers & Cars & Pets & CUB & Average   \\
		
		\midrule
        Fine-tuning-Ours &1, 1, 1 & 78.03&	57.47&	79.71&	38.27&	71.19&	39.64&	60.72\\
        Training-from-scratch-Ours &1, 1, 1 &42.57&	26.96&	5.88&		1.06&	5.12&	1.64& 13.87\\
        Training-from-scratch-PSQ &1, 1, 1 & 24.27&	6.14&	2.94&	1.04&	4.41&	1.04&	6.64\\

        \bottomrule
        \end{tabular}
        \label{scratch}
    \end{table*}

\begin{figure}[t]
\vskip -0.1in
\centering
\begin{minipage}[b]{0.48\textwidth}
  \centering
  \includegraphics[width=0.935\textwidth]{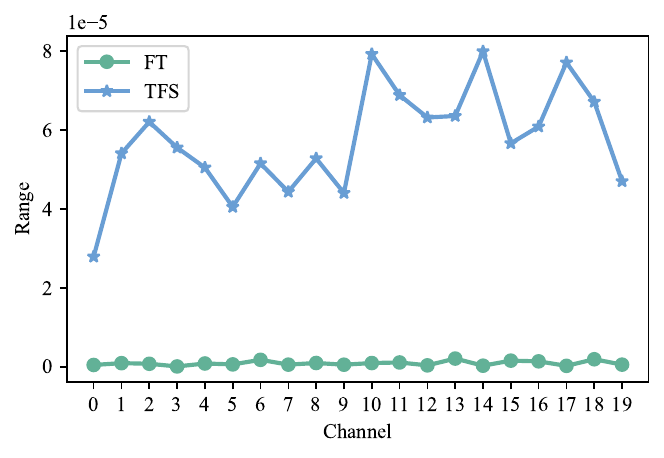}
  \label{gradient:left}
  
\end{minipage}
\hfill
\begin{minipage}[b]{0.48\textwidth}
  \centering
  \includegraphics[width=1\textwidth]{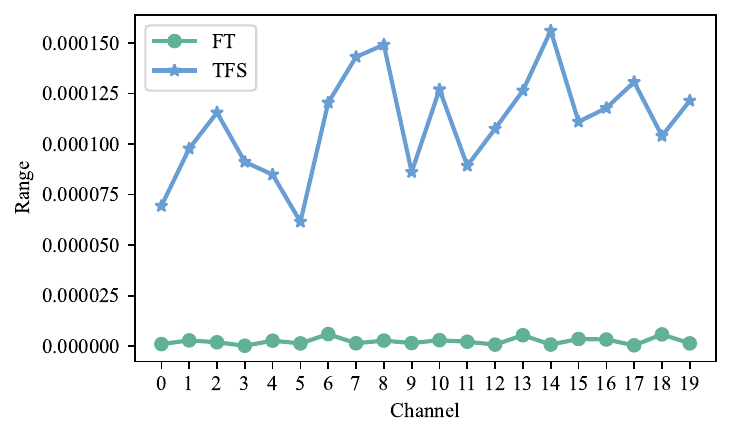}
  \label{gradient:right}
  
\end{minipage}
\caption{Gradient range analysis in ResNet18's conv2.1.2 layer under fine-tuning (FT) and training from scratch (TFS). (left) The result from CIFAR-10. (right) The result from CIFAR-100.}
\vskip -0.2in
\label{gradient}
\end{figure}
\subsection{FQT from scratch}
We compared the performance of our method in two scenarios: fine-tuning and training from scratch. We presented the classification results under the optimal configuration (b=4) in Table \ref{scratch}. From the table, it is evident that when training from scratch, the model exhibits very low classification accuracy across all datasets, and in certain datasets, it even lacks classification capability entirely. We attempted to analyze the differences between the two scenarios based on the distinct gradient distributions. From Fig. \ref{gradient}, we observe that the gradient range is larger in training from scratch, leading to increased gradient variance (Eq. \ref{eq24}) and reduced model convergence. Therefore, 1-bit FQT from scratch remains an open problem. Additionally, we also compared our method with PSQ in the scenario of training from scratch, and the results indicate that our approach still significantly outperforms PSQ in accuracy.

\begin{figure}[t]
  \centering

    \subfigure[]{\includegraphics[width=0.45\columnwidth]{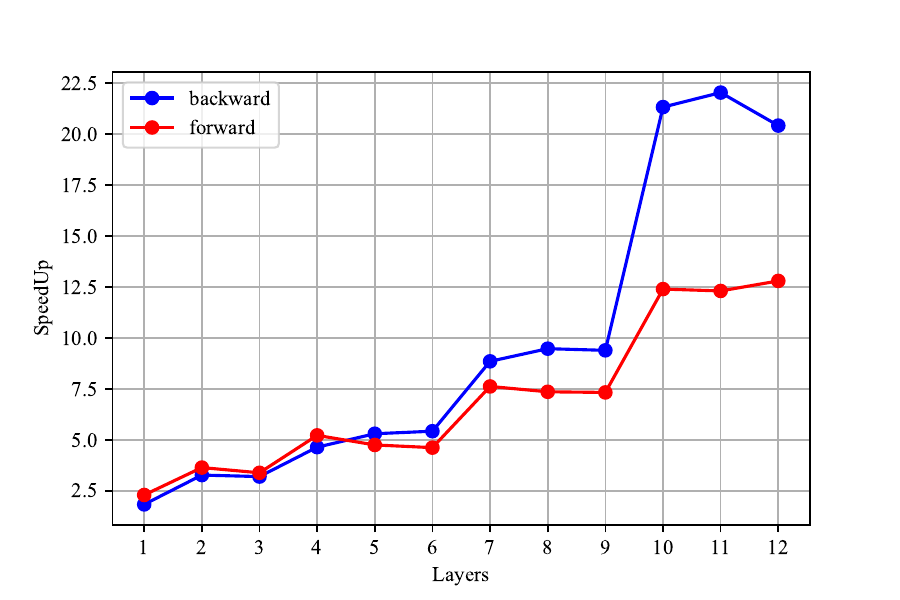}
    \label{fig5:left}}
    \subfigure[]{\includegraphics[width=0.45\columnwidth]{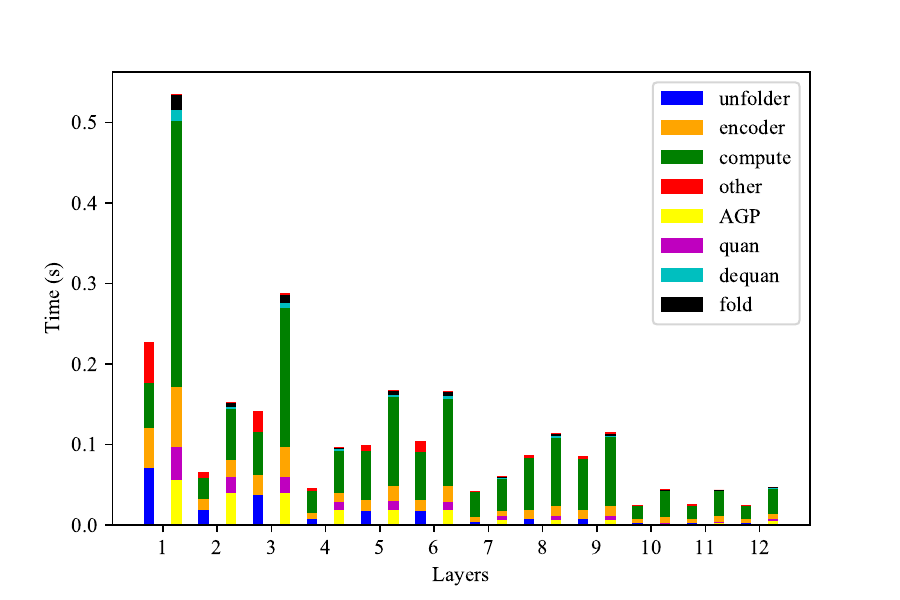}
    \label{fig5:right}}
 
  \caption{(a) The speedup of ours compared with FP32 PyTorch. (b) The compositional structure of time consumption.}
  \label{fig5}
\end{figure}
\subsection{Time expenditure structure}
We present the speedup across layers of VGGNet16 and the
time consumption for each operation in Fig. \ref{fig5}, providing guidance
for future optimization directions. It is important to note that the first and last layers were not quantized and, therefore, were not included in the analysis. From the figure, it is evident that matrix multiplication constitutes the majority of the training time, while the time overhead of other operations such as gradient pruning and quantization can be considered negligible. Therefore, the focus of future optimization efforts will remain on matrix multiplication. Furthermore, it can be observed that our implemented method is particularly friendly for layers with a large number of convolutional kernels and smaller input resolution.


\newpage
\section*{NeurIPS Paper Checklist}



\begin{enumerate}

\item {\bf Claims}
    \item[] Question: Do the main claims made in the abstract and introduction accurately reflect the paper's contributions and scope?
    \item[] Answer: \answerYes{} 
    \item[] Justification: In Sec. \ref{introduction}, we state our contributions along with the scope and limitations of our method.

\item {\bf Limitations}
    \item[] Question: Does the paper discuss the limitations of the work performed by the authors?
    \item[] Answer: \answerYes{} 
    \item[] Justification: In Sec. \ref{introduction} and \ref{conclusion}, we discuss the limitations of this study.

\item {\bf Theory Assumptions and Proofs}
    \item[] Question: For each theoretical result, does the paper provide the full set of assumptions and a complete (and correct) proof?
    \item[] Answer: \answerYes{} 
    \item[] Justification: In Sec. \ref{theory}, theoretical results, including assumptions, are presented, with proofs provided in Appendix \ref{appendix:proof}.

    \item {\bf Experimental Result Reproducibility}
    \item[] Question: Does the paper fully disclose all the information needed to reproduce the main experimental results of the paper to the extent that it affects the main claims and/or conclusions of the paper (regardless of whether the code and data are provided or not)?
    \item[] Answer: \answerYes{} 
    \item[] Justification: In the supplementary materials, we offer the code to accomplish this.

\item {\bf Open access to data and code}
    \item[] Question: Does the paper provide open access to the data and code, with sufficient instructions to faithfully reproduce the main experimental results, as described in supplemental material?
    \item[] Answer: \answerYes{} 
    \item[] Justification: In the supplementary materials, we offer the code.

\item {\bf Experimental Setting/Details}
    \item[] Question: Does the paper specify all the training and test details (e.g., data splits, hyperparameters, how they were chosen, type of optimizer, etc.) necessary to understand the results?
    \item[] Answer: \answerYes{} 
    \item[] Justification: In Sec. \ref{Experiment}, we outline the core experimental setup, with the complete setup detailed in Appendix \ref{appendix:experiment}.

\item {\bf Experiment Statistical Significance}
    \item[] Question: Does the paper report error bars suitably and correctly defined or other appropriate information about the statistical significance of the experiments?
    \item[] Answer: \answerYes{} 
    \item[] Justification: We report the mean and stddev of 3 runs for main results (visual classification task).

\item {\bf Experiments Compute Resources}
    \item[] Question: For each experiment, does the paper provide sufficient information on the computer resources (type of compute workers, memory, time of execution) needed to reproduce the experiments?
    \item[] Answer: \answerYes{} 
    \item[] Justification: We describe the type and model of compute workers in Sec. \ref{Experiment} and report the time of execution.
    
\item {\bf Code Of Ethics}
    \item[] Question: Does the research conducted in the paper conform, in every respect, with the NeurIPS Code of Ethics \url{https://neurips.cc/public/EthicsGuidelines}?
    \item[] Answer: \answerYes{} 
    \item[] Justification: I confirm that the paper complies with ethical guidelines in every aspect.

\item {\bf Broader Impacts}
    \item[] Question: Does the paper discuss both potential positive societal impacts and negative societal impacts of the work performed?
    \item[] Answer: \answerYes{} 
    \item[] Justification: In the discussion following the conclusion, we explored the potential positive impacts of our method; since our approach aims to accelerate network training, no significant negative effects were observed.
    
\item {\bf Safeguards}
    \item[] Question: Does the paper describe safeguards that have been put in place for responsible release of data or models that have a high risk for misuse (e.g., pretrained language models, image generators, or scraped datasets)?
    \item[] Answer: \answerNA{} 
    \item[] Justification: Since this paper focuses on accelerating network training, it does not present these risks.

\item {\bf Licenses for existing assets}
    \item[] Question: Are the creators or original owners of assets (e.g., code, data, models), used in the paper, properly credited and are the license and terms of use explicitly mentioned and properly respected?
    \item[] Answer: \answerYes{} 
    \item[] Justification: The assets used in this paper are credited and the license is respected.

\item {\bf New Assets}
    \item[] Question: Are new assets introduced in the paper well documented and is the documentation provided alongside the assets?
    \item[] Answer: \answerYes{} 
    \item[] Justification: New assets introduced in the paper are well documented.

\item {\bf Crowdsourcing and Research with Human Subjects}
    \item[] Question: For crowdsourcing experiments and research with human subjects, does the paper include the full text of instructions given to participants and screenshots, if applicable, as well as details about compensation (if any)? 
    \item[] Answer: \answerNA{} 
    \item[] Justification: The paper does not involve crowdsourcing or research with human subjects.

\item {\bf Institutional Review Board (IRB) Approvals or Equivalent for Research with Human Subjects}
    \item[] Question: Does the paper describe potential risks incurred by study participants, whether such risks were disclosed to the subjects, and whether Institutional Review Board (IRB) approvals (or an equivalent approval/review based on the requirements of your country or institution) were obtained?
    \item[] Answer: \answerNA{} 
    \item[] Justification:  The paper does not involve crowdsourcing nor research with human subject.

\end{enumerate}

\end{document}